%% file: bare_jrnl.tex
\documentclass[journal]{IEEEtran}
\usepackage{amsmath,amssymb}
\usepackage{graphicx}
\usepackage{multirow,booktabs}
\usepackage{algorithm}
\usepackage{algorithmic}
\usepackage{cite}
\usepackage{xcolor}
\usepackage{soul}
\usepackage[normalem]{ulem}
\usepackage{subcaption}
\usepackage[hidelinks]{hyperref}

\input{my_acronyms}
\input{my_definitions}

\begin{document}
%

\title{Coupled Control and Wireless World Models for Resilient Remote Robotic Control}

%


\author{
\IEEEauthorblockN{
H.P.~Madushanka,~\IEEEmembership{Student Member,~IEEE,}
Sumudu~Samarakoon,~\IEEEmembership{Member,~IEEE,} and
Mehdi Bennis,~\IEEEmembership{Fellow,~IEEE}
}


%
%
\thanks{The authors are with the Centre for Wireless Communications, University of
Oulu, 90570 Oulu, Finland
(e-mail: madushanka.hewapathiranage@oulu.fi;
sumudu.samarakoon@oulu.fi;
mehdi.bennis@oulu.fi).}
}

\maketitle

\instructions{Use the following rules:\\
- All abbreviations should be used with package glossary. Check the RRT definition in my\_acronyms.tex. Use the same format for all. Make sure to use correct command when necessary gls, Gls, glspl, Glspl \\
- Define all symbols and other commands/notations in my\_definitions.tex. Check the example left under the latter part of the Introduction section. \\
- Do not overuse latex packages. Try to stick with the minimum, only add new packages if the existing ones cannot handle the desired requirements. \\
}

\begin{abstract}
Remote robotic systems operating over wireless networks must maintain reliable control despite limited communication resources, changing channel conditions, and environmental disturbances.
However, continuously transmitting high-dimensional sensory observations, such as camera images, increases communication overhead and energy consumption while reducing robustness under unreliable connectivity.
To address these challenges, this paper proposes a resilient communication-aware remote robotic control framework based on coupled control and wireless \gls{jepa} world models that jointly capture robot dynamics and wireless channel evolution from visual observations and a combination of raw and structured \gls{rf} representations based on spectrograms and \glspl{pi}.
The learned latent representations enable predictive communication scheduling by jointly forecasting future robot states and wireless conditions, thereby reducing unnecessary uplink transmissions while maintaining reliable control performance.
Furthermore, an adaptive resilience mechanism detects latent prediction discrepancies and efficiently adapts perception embeddings to accommodate wireless and visual environmental changes without retraining the complete control policy.
The proposed framework is evaluated in a synchronized Gazebo--\gls{ros}--Sionna robot--wireless simulation environment under diverse wireless propagation and perception perturbations.
Experimental results demonstrate significant improvements in communication efficiency, robustness, and resilience while maintaining navigation performance compared with conventional \gls{pid}, model-free \gls{dqn}, and predictive approaches based on \glspl{vit}.

\end{abstract}


%
\IEEEpeerreviewmaketitle

\section{Introduction}

\instructions{
Introduce the remote control/operation over limited/stochastic connectivity. 
Motivate with some examples and references.

Next we need to introduce \glspl{wm}. 
Discuss the relation to \glspl{dt}, transition models used in model-based learning, statistical models.
For what they are used? How do we utilize them towards resiliency?
Also, we can specify the need for multiple \glspl{wm} that rely on different modalities.

Introduce JEPA.
Refer Charbel's paper and discuss about the C-JEPA and W-JEPA.

Clearly state the contributions:

\begin{itemize}
    \item The realistic sandbox environment (? Add this unless we plan for a separate paper for SIONNA-GAZEBO integration)
    \item Transfer learning from rl-gym model to Gazebo as a remedy for limited computing
    \item The use of different modalities as well as their representations (Image and RF, RF as spectrograms as well as PDs) 
    \item Designs and investigations on robustness and resilience
\end{itemize}

Outline the organization of the paper.

\tred{
Example reference \cite{lavalle2001rapidly}.
}

\tred{
The state of the agent is $\state\in\stateSpace$ where $\stateSpace$ is the state space. 
}
------------------------------------------------
}

\glsresetall

Remote robotic systems are becoming increasingly important in applications such as warehouse automation, industrial inspection, disaster response, healthcare, autonomous transportation, and smart manufacturing \cite{warehouse_logistics,future_robot,darvish2023teleoperation}.
Therein, the connectivity between the robots and edge servers/centralized controllers for sensing and controlling mostly relies on wireless communication links, and future intelligent robotic systems are expected to leverage emerging \mbox{6G} networks to provide ultra-reliable, low-latency, and AI-native communication services for distributed autonomy and edge intelligence \cite{Park2022Extreme,8869705}.
Continuously transmitting high-dimensional sensory observations, such as camera images, depth maps, or multimodal sensor data, can significantly increase communication overhead, latency, and energy consumption 
thereby reducing communication efficiency \cite{Park2022Extreme,10012674}.
Furthermore, wireless channel variations, intermittent packet losses, and environmental disturbances can considerably degrade control performance in remote robotic systems \cite{5984917,4758193}.
Hence, it is crucial to maintain reliable remote operation under limited communication resources, stochastic wireless connectivity, and dynamic environmental conditions.

Recent advances in \gls{ai} and model-based learning have motivated the development of predictive frameworks capable of learning system dynamics directly from sensory observations.
Among these approaches,
\glspl{wm} that characterize system dynamics have emerged as a promising framework for learning compact latent representations of environments and predicting future observations and states \cite{wm,hafner2019learning}.
\Glspl{wm} are widely used in model-based \gls{rl}, autonomous driving, embodied intelligence, and robotic planning systems because they allow agents to reason about future system behavior instead of relying solely on reactive control policies \cite{hafner2023mastering,wu2023daydreamer}. 
%
Additionally, \glspl{wm} are becoming increasingly relevant in realistic simulation frameworks and digital twins, where high-fidelity virtual environments are used to replicate physical systems while their predictive capabilities are used to support intelligent control and optimization in realistic settings \cite{Grieves2017}.
%
%
%
%
%
%
Reliable remote robotic control with \glspl{wm} requires predictive understanding of both the physical environment, which governs robot behavior, and the wireless communication environment that determines the quality of information exchange between the robot and the remote controller\cite{11480198}.
%
%
Jointly modeling these complementary dynamics therefore motivates the development of multimodal \glspl{wm} that learn predictive latent representations from visual observations and wireless communication measurements.

To effectively model complex system dynamics, \glspl{wm} require informative latent representations that capture the underlying structure of the environment and enable accurate prediction of future system states. Among recent predictive representation learning approaches, \glspl{jepa} have emerged as a promising framework for learning such predictive latent representations directly from sensory observations \cite{lecun2022path}.
%
Unlike conventional generative models operating directly in pixel space, \gls{jepa}s focus on learning semantically meaningful predictive embeddings that are better suited for downstream control and prediction tasks.
Recent studies have demonstrated the effectiveness of \glspl{jepa} in visual representation learning, video understanding, and embodied intelligence, highlighting their ability to learn predictive representations that generalize across diverse environments and tasks \cite{lecun2022path,assran2023self,bardes2024revisiting}.
Building upon these advances, a coupled \gls{cjepa} and \gls{wjepa} framework was proposed in \cite{chaaya2025pixels} for jointly modeling control and wireless dynamics in a latent space to optimize wireless resource utilization while maintaining control performance in remote robotic systems.
In particular, the framework combined latent control dynamics learned from visual observations with wireless \gls{csi} dynamics through cross-modal conditioning.
Their results demonstrated reductions in communication overhead and transmit power while maintaining comparable control performance to model-free \gls{rl} approaches.
Yet the above framework has primarily been validated in simplified simulation environments and does not sufficiently investigate robustness and resilience under realistic wireless and environmental disturbances.
Furthermore, transferring predictive \glspl{wm} from lightweight \gls{rl} environments to realistic robotic simulators remains a challenging problem due to differences in sensing, environment dynamics, and computational complexity.
In addition, the use of predictive wireless representations within \glspl{wm} to enable resilient communication-aware robotic control under dynamic wireless environments remains relatively underexplored \cite{comunication_aware,li2025robotic,chaaya2025pixels}.

Motivated by these challenges, this paper extends the coupled \gls{cjepa} and \gls{wjepa}  framework proposed in \cite{chaaya2025pixels} by developing a realistic communication-aware robotic control framework within a simulation environment integrating Gazebo, \gls{ros}, and the Sionna \gls{rt} simulator \cite{gazebo,hoydis2022sionna,quigley2009ros}.
Unlike the original framework, the proposed design addresses the challenges of deploying predictive \glspl{wm} in a realistic robotic simulation environment characterized by richer sensing, more complex environmental dynamics, and increased computational requirements.
Towards this end, a transfer learning strategy is introduced to efficiently adapt a pretrained control \gls{wm} from the Gym environment to Gazebo, thereby reducing the computational cost and training time associated with learning directly in realistic robotic environments.
Furthermore, to better understand the impact of wireless representations on predictive wireless world modeling, this work systematically evaluates multiple wireless representations, including raw wireless channel measurements and structured \gls{rf} representations.
To enable resilience under dynamic environmental conditions, an adaptive perception mechanism is further proposed to detect discrepancies between predicted and observed robot perceptions and trigger appropriate observation adaptation in response to environmental changes.
Finally, the proposed framework is extensively evaluated under both wireless communication disturbances and perception disturbances, demonstrating its robustness to communication impairments and its resilience through adaptive recovery under changing operating conditions.
%
%
The main contributions of this paper are summarized as follows:
\begin{itemize}
    \item We develop a realistic communication-aware synchronized robot-wireless simulation framework by integrating Gazebo, \gls{ros}, and the Sionna \gls{rt} simulator, enabling joint robotic control and wireless communication experiments under realistic operating conditions.
    \item We extend the coupled \gls{cjepa} and \gls{wjepa} framework proposed in \cite{chaaya2025pixels} from a lightweight \gls{rl} Gym environment to a realistic simulation environment through an efficient transfer learning strategy, reducing the computational cost and training effort required for predictive world modeling in complex robotic environments.
    \item We comprehensively evaluate the robustness of the proposed framework under diverse wireless communication disturbances, including carrier-frequency shifts, scatterer variations, access-point handover, and intermittent uplink communication failures.
    \item We propose an adaptive resilience mechanism that detects discrepancies between predicted and observed robot perceptions and updates the perception pipeline, enabling recovery from environmental changes during remote robotic operation.
    \item We systematically investigate multiple wireless representations, including raw wireless channel measurements and structured \gls{rf} representations, to analyze their effectiveness for predictive wireless world modeling and communication-aware remote robotic control.
\end{itemize}

The rest of the paper is organized as follows.
The related work is first summarized in Sec.~\ref{sec:sota}.
Sec.~\ref{sec:sys_mod} presents the system architecture and implementation related details.
The experimental settings and performance evaluation are discussed in Sec.~\ref{sec:results}.
Finally, Sec.~\ref{sec:conclusion} concludes the paper and outlines future research directions.

\section{Related Work}\label{sec:sota}

\instructions{
\sps{Note}{We do not need subsections here. Just organize the SOTA under the topics as paragraphs}

\st{
-- World models and JEPA

-- Control and communication designs

-- Robustness and resilience

-- Realistic simulations (Gazebo, Sionna, and some that are close to their joint designs)
}
-------------------------------------------
}

Recent advances in model-based learning have established \glspl{wm} as a promising framework for learning compact latent representations of complex environments and predicting future system dynamics.
By learning predictive representations directly from sensory observations, \glspl{wm} enable planning, decision making, and control while reducing the need for extensive real-world interactions \cite{wm,hafner2019learning,hafner2023mastering,wu2023daydreamer}.
Such approaches have demonstrated strong performance in \gls{rl} \cite{hafner2019learning,hafner2023mastering,micheli2022transformers}, autonomous navigation \cite{wu2023daydreamer,li2025robotic}, and embodied intelligence \cite{bardes2024revisiting,assran2025v}, where latent representations are used to forecast future states and support long-horizon reasoning.
More recently, predictive representation learning methods have shifted attention from observation reconstruction toward latent prediction, allowing models to focus on semantically meaningful information while improving learning efficiency \cite{lecun2022path,assran2023self,assran2025v}.
These developments have motivated the use of latent predictive models for robotics, communication systems, and autonomous decision making.
However, most existing world-model frameworks primarily focus on physical environment dynamics and do not explicitly account for wireless communication conditions or communication-resource constraints.

The integration of control and communication has attracted increasing attention in wireless robotic systems and networked control applications.
Traditional approaches focus on maintaining stability and control performance under communication limitations through scheduling, estimation, and resource allocation mechanisms \cite{VR,5984917}. 
More recent studies have explored communication-aware control strategies that jointly optimize communication and control decisions to improve overall system performance \cite{VR,comunication_aware,9493202}. 
Building upon these ideas, predictive latent models have recently been extended to jointly capture control and wireless dynamics, enabling communication-aware decision making and wireless resource management through latent-space reasoning \cite{chaaya2025pixels,9493202}.
Despite these advances, the application of communication-aware latent \glspl{wm} to realistic robotic environments remains relatively unexplored.

Robotic systems operating in real-world environments must remain robust to uncertainties arising from communication failures, sensing errors, environmental changes, and model inaccuracies.
Existing research has investigated robustness through adaptive control, probabilistic estimation, sensor fusion, and uncertainty-aware learning techniques \cite{article_kober,6792214}.
Similarly, resilience in wireless systems has been addressed through channel estimation, adaptive scheduling, diversity techniques, and robust communication protocols \cite{10012674,9252948}.
While these approaches improve performance under specific disturbances, they often treat communication and control as independent problems.
Furthermore, most existing methods focus on maintaining performance under known disturbances rather than adapting to previously unseen environmental changes.
The use of predictive latent models for jointly identifying communication anomalies, perception mismatches, and environmental changes remains an emerging research direction.

The increasing complexity of intelligent robotic systems has also highlighted the importance of realistic  and high-fidelity simulation environments.
High-fidelity simulation environments provide a safe and cost-effective platform for developing, testing, and validating autonomous systems prior to real-world deployment \cite{Grieves2017,TAO2019653}.
Within robotics, Gazebo has become one of the most widely adopted simulation platforms due to its realistic physics engine, sensor models, and integration with the \gls{ros} \cite{gazebo,quigley2009ros}.
In wireless communications, Sionna has recently emerged as a powerful framework for ray-tracing-based wireless simulation, enabling realistic modeling of propagation effects, channel dynamics, and communication performance \cite{hoydis2022sionna}.
Although both platforms have demonstrated significant value independently, relatively few studies have investigated their joint integration for communication-aware robotic systems.
Existing robotic simulations often rely on simplified wireless assumptions, while wireless simulations typically abstract away robot dynamics and perception. 
As a result, the interaction between robot control, wireless communication, and environmental changes remains insufficiently explored within unified synchronized robot–wireless simulation environments.

\section{System Architecture, Methodology, and Implementation}\label{sec:sys_mod}

\begin{figure}[t]
    \centering
    \includegraphics[width=\linewidth]{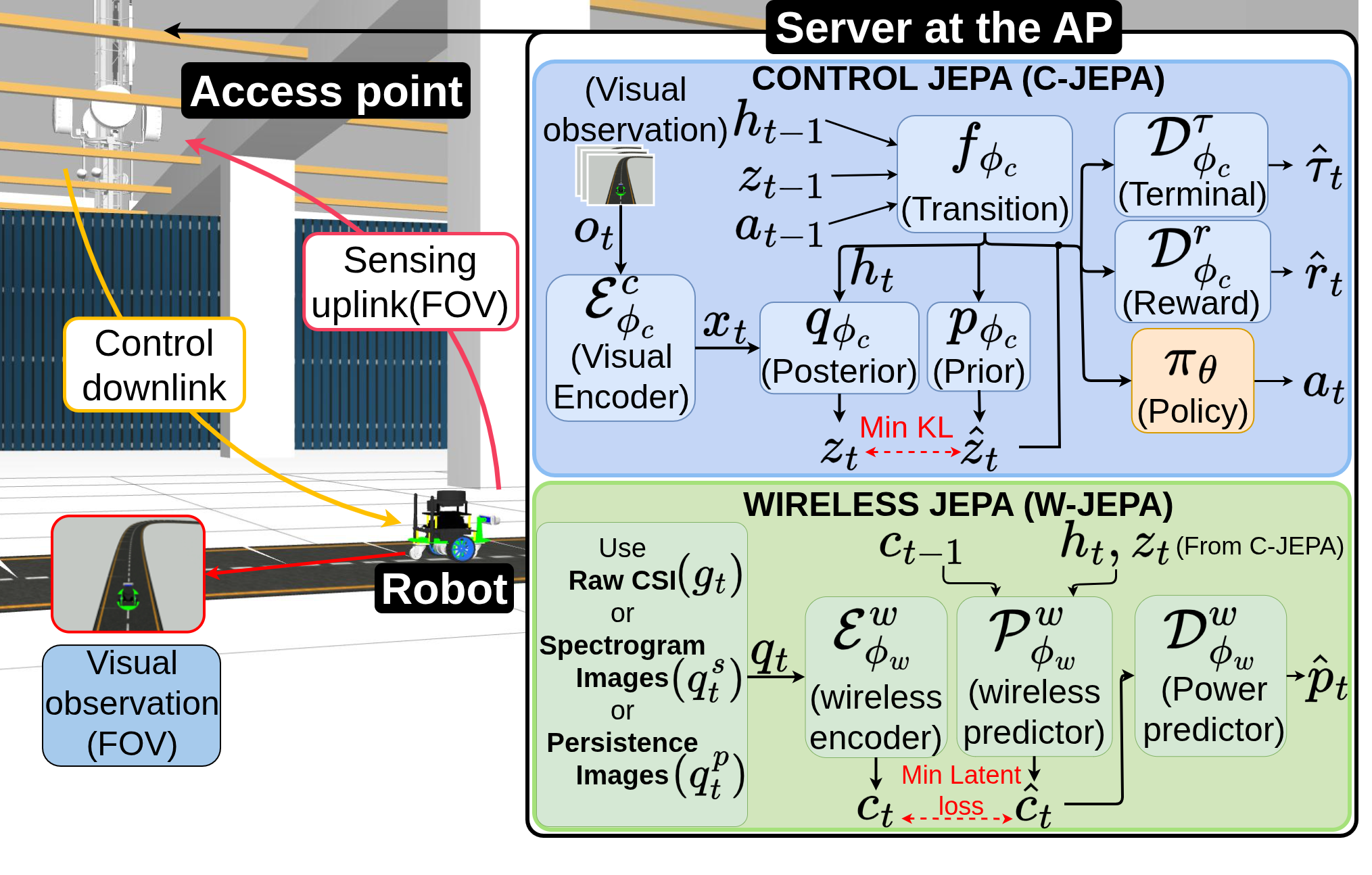}
    \caption{Proposed communication-aware remote robotic control framework based on coupled \gls{cjepa} and \gls{wjepa}.\\
    }
    \label{fig:overall_figure}
\end{figure}

%
%
We consider a system consists of a mobile robot, a wireless communication infrastructure, and a remote controller connected through a wireless \gls{ap} as illustrated in Fig. \ref{fig:overall_figure}.
%
%
%
The robot 
uses an onboard sensing system to acquire visual observations $\vect{o}_t$ at each time step $t$, and transmits them to the remote controller through the uplink channel.
Simultaneously, 
the uplink \gls{csi} $\vect{g}_t$ or alternative wireless representations derived from \gls{rf} measurements $\vect{q}_t$ are used at the controller as wireless observations.
Together with the visual and wireless observations, the controller derives the control commands $a_t$ and sends them to the robot through the downlink channel, forming a closed-loop control architecture.
Based on the received control commands, the robot performs a navigation task.

The physical environment consisting of solid objects, light sources, texture, and terrain serves multiple purposes. 
It depicts the world that the robot has to interact and operate, in which, lighting and texture with objects create realistic visual sensory observations while the terrain and obstacles together defines the robot's mobility dynamics.
Additionally, the objects with different \gls{rf} materials acting as scatters along with the robots movements directly impact on the wireless channel dynamics. 
For such a realistic depiction of the wireless conditions, we rely on the Sionna \gls{rt} simulator to model both the uplink and downlink channels using physics-based \gls{rt} over the synchronized scene geometry, where material-dependent propagation is based on the \gls{itu}-defined radio material models that captures \gls{los} and multipath propagation effects.
Unlike the remote controller, the robot is constrained with a limited energy, and thus, its uplink communication is scheduled by the controller towards the utilization of wireless resources \cite{9991044}. 

In this view, a binary scheduling variable $\rho_t$ is introduced, where $\rho_t=1$ indicates the uplink channel use for transmitting an observation while $\rho_t=0$ indicates that the transmission is omitted.
It is worth noting that for scheduling, the controller needs to be aware of future uplink channel conditions, i.e., wireless dynamics, while controlling with the cases of $\rho_t=0$, the controller needs to be aware of future observations, i.e., sensing and control dynamics.
Hence, we consider that the controller is equipped with \glspl{wm} that allow predicting wireless and sensing observations through a series of learned latent representations.
Together,
the framework pursues four main objectives:
\begin{itemize}
    \item Maintain reliable robotic control and task performance under limited wireless connectivity.
    \item Reduce communication overhead through predictive transmission of informative observations and compact latent representations.
    \item Improve wireless resource utilization by exploiting favorable predicted channel conditions.
    \item Enhance resilience via adapting to dynamic wireless and environmental changes that may affect communication and control performance.
\end{itemize}

Towards achieving these objectives, the proposed framework employs two coupled predictive \glspl{wm}, namely the \gls{cjepa} and \gls{wjepa}, to jointly model the dynamics of the robot movements and wireless conditions within a unified latent representation space.
Here, the \gls{cjepa} learns predictive representations of the robot and its environment from visual observations, while the \gls{wjepa} learns the evolution of wireless communication conditions from \gls{csi} or \gls{rf}-derived spectrogram observations.
The latent representations generated by both \gls{jepa} models are subsequently utilized for control policy learning, wireless power prediction, and predictive communication scheduling.
By jointly reasoning about future robot behavior and future communication conditions, the framework enables communication-aware remote robotic control under limited and dynamic wireless connectivity.

\subsection{Joint \gls{cjepa} and \gls{wjepa} Learning}

As illustrated in Fig.~\ref{fig:overall_figure}, the proposed framework consists of two coupled predictive models, namely the \gls{cjepa} and \gls{wjepa}, following the JEPA-based predictive world modeling framework presented in \cite{chaaya2025pixels}. 
The \gls{cjepa} and \gls{wjepa} are parameterized by the learnable weights $\phi_c$ and $\phi_w$, respectively, and jointly optimized through their corresponding predictive learning objectives. 
During training, the \gls{cjepa} first learns predictive latent control dynamics from visual observations, while the \gls{wjepa} subsequently learns predictive wireless dynamics by encoding wireless observations and conditioning the latent wireless prediction on the predictive control latent representations generated by the \gls{cjepa}.
The joint learning process is described as follows.

Once the robot acquires a visual observation $\vect{o_t}$, it is transmitted to the remote controller through the sensing uplink and encoded into a compact latent representation
$
x_t = \mathcal{E}_{\phi_c}^{c}(\vect{o_t})
$
where $\mathcal{E}_{\phi_c}^{c}(\cdot)$ denotes the visual encoder and  represents the latent visual embedding.
The encoded representation is subsequently propagated through a \gls{rssm}, which learns the temporal evolution of the robot dynamics using a deterministic latent state $h_t$ and a stochastic latent state $z_t$. 
The deterministic latent state captures the temporal memory of the system, while the stochastic latent state represents the uncertainty of the underlying robot dynamics.

The \gls{rssm} consists of three learnable components: the recurrent transition model $f_{\phi_c}(\cdot)$, the prior model $p_{\phi_c}$, and the posterior model $q_{\phi_c}$.
The recurrent transition model first updates the deterministic latent state according to
$
h_t=f_{\phi_c}(h_{t-1},a_{t-1},z_{t-1}) 
$,
where $a_{t-1}$ denotes the previous control action.
Conditioned on the updated $h_t$, the prior model predicts the stochastic latent representation as
$
\hat{z}_t \sim p_{\phi_c}(\hat{z}_t \mid h_t)
$,
while the posterior model incorporates the encoded visual representation to infer the observation-conditioned latent state according to
$
z_t \sim q_{\phi_c}(z_t \mid h_t,x_t)
$.

The learned latent representations $h_t$ and $z_t$ are subsequently utilized by the reward prediction model $\mathcal{D}_{\tau}^c(\cdot)$, the terminal prediction model $\mathcal{D}_{r}^c(\cdot)$, and the control policy model $\pi_{\theta_c}(\cdot)$, parameterized by the learnable weights $\theta_c$.  
Specifically, the reward and terminal prediction models estimate the immediate reward and episode termination probability, respectively,
$
\hat{r}_t = \mathcal{D}^{c}_{r}(h_t,z_t)
$,
$
\hat{\tau}_t = \mathcal{D}^{c}_{\tau}(h_t, z_t)
$,
while the control policy generates the corresponding robot control action according to
$
a_t = \pi_{\theta_c}(h_t, z_t)
$.

During training, the \gls{cjepa} is optimized over sequences of length $T$ sampled from the collected experience, with the \gls{wm} objective accumulated over the sequence as
%
$\phi_c^{*}
=
\arg\min_{\phi_c}
\mathcal{L}_{C}(\phi_c)$
%
where
\begin{equation}
\begin{aligned}
\mathcal{L}_{C}(\phi_c)
=
\mathbb{E}_{q_{\phi_c}}
\Bigg[
\sum_{t=1}^{T}
\Big(
&\beta
D_{\mathrm{KL}}
\left(
z_t
\,\|\, 
\hat{z_t}
\right) 
-
\log p_{\phi_c}(r_t \mid \hat{r}_t) \\
&-
\log p_{\phi_c}(\tau_t \mid \hat{\tau}_t
\Big)
\Bigg].
\end{aligned}
\label{eq:cjepa_loss}
\end{equation}
Here, $D_{\mathrm{KL}}(\cdot)$ denotes the \gls{kl} divergence between the posterior and prior latent distributions, while the second and third terms correspond to the reward prediction and terminal prediction log-likelihood losses, respectively.

%

In parallel, the \gls{wjepa} receives the wireless observation $q_t$, which represents a generic wireless observation. 
Depending on the selected \gls{rf} representation, $q_t$ may correspond to the raw \gls{csi}, spectrogram image, or \gls{pi}, as described in the following subsection. 
The wireless observation is first encoded into a compact latent representation according to
$
c_t=\mathcal{E}_{\phi_w}^w(q_t)
$,
where $\mathcal{E}_{\phi_w}^w(\cdot)$ denotes the wireless encoder and $c_t$ is the corresponding latent wireless representation.
The encoded representation is subsequently processed by the wireless predictor $\mathcal{P}_{\phi_w}^w(\cdot)$, which predicts the future wireless latent representation by jointly exploiting the previous wireless latent representation and the predictive control latent states according to
$
\hat{c}_t = \mathcal{P}_{\phi_w}^w(c_{t-1},h_t,z_t)
$.

Similarly, the wireless encoder and wireless predictor are trained by minimizing the discrepancy between the predicted wireless latent representation $\hat{c}_t$ and the encoded target representation $c_t$ over a training sequence of length $T$.
The corresponding latent prediction loss is
\begin{equation}
\mathcal{L}_{W}(\phi_{w})
=
\mathbb{E}_{q_{\phi_w}}
\Big[
\sum_{t=1}^{T}
\left\|
\hat{c}_{t}-c_{t}
\right\|_{2}^{2}
\Big].
\label{eq:wjepa_loss}
\end{equation}
This objective encourages the \gls{wjepa} to learn temporally consistent latent representations of the wireless dynamics.

The predicted wireless latent representation is subsequently processed by the wireless power prediction model $\mathcal{D}_{\phi_w}^w(\cdot)$ to estimate the future uplink transmission power according to 
$
\hat{p}_t = \mathcal{D}_{\phi_w}^w(\hat{c}_t).
$
The power prediction model is trained by minimizing the discrepancy between the predicted transmission power $\hat{p}_t$ and the corresponding target power $p_t$ given by,
\begin{equation}
\mathcal{L}_{P}(\phi_{w})
=
\sum_{t=1}^{T}
\left|
\hat{p}_{t}-p_{t}
\right|.
\label{eq:power_loss}
\end{equation}
The predicted transmission power $\hat{p}_t$ is subsequently utilized for communication-aware uplink scheduling.
Based on the resulting scheduling decision $\rho_t$, the current visual observation is transmitted to the remote controller or the controller continues predicting the robot state using the learned \gls{cjepa} latent dynamics.

\begin{figure}
    \centering
    \includegraphics[width=\linewidth]{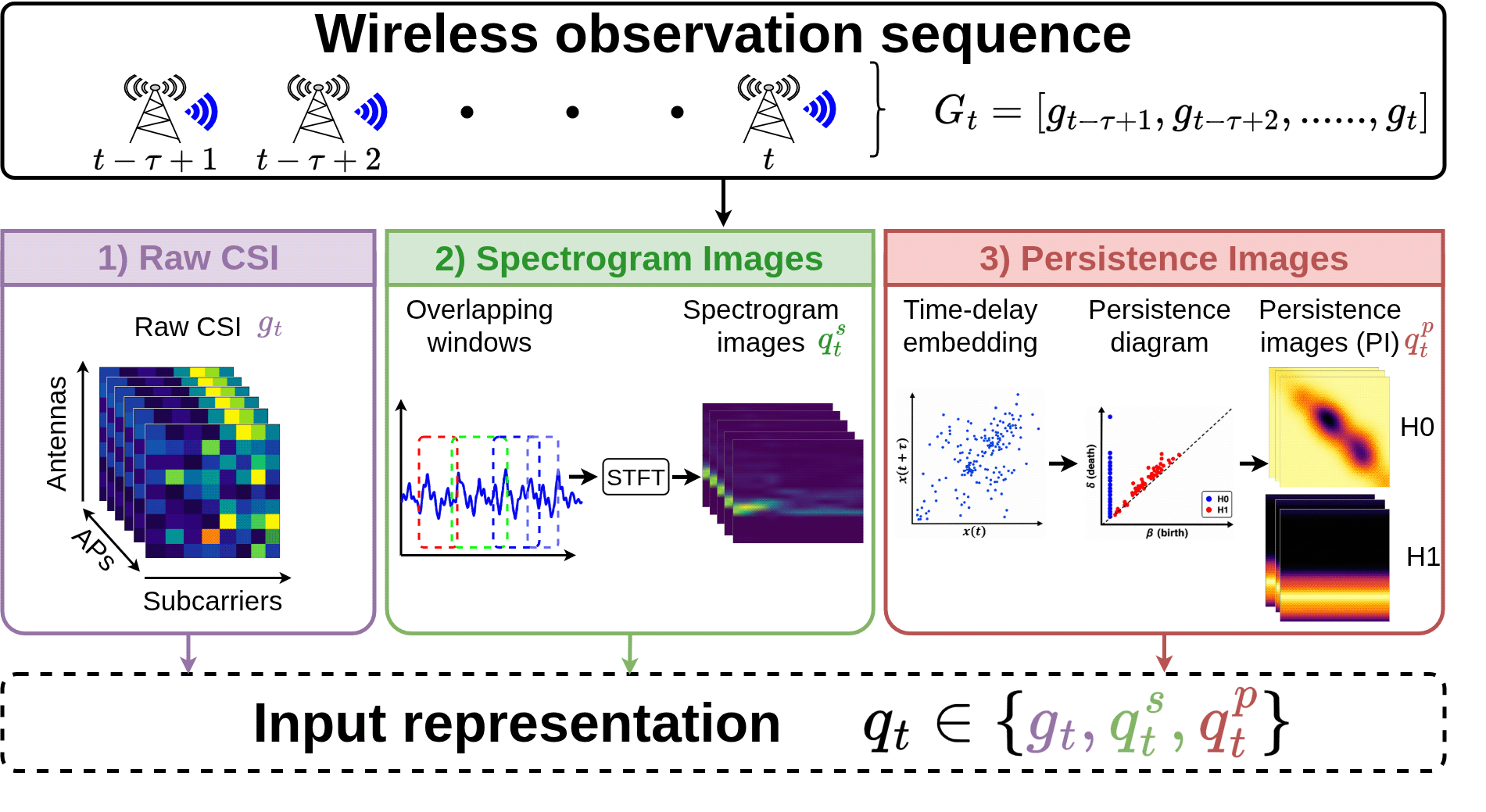}
    \caption{Structured \gls{rf} representation generation from the wireless observation sequence $\vect{G}_t$. Three alternative input representations are considered for the \gls{wjepa}: raw \gls{csi} $\vect{g}_t$, spectrogram images $\vect{q}_t^s$, and \glspl{pi} $\vect{q}_t^p$.
    }
    \label{fig:wireless_embeddings}
\end{figure}

\subsection{Structured \gls{rf} Representations for \gls{wjepa}} \label{sec:wireless_embeddings}

%
Towards improving the identification of temporal changes in wireless conditions, we further study the use of \gls{csi}-derived structured \gls{rf} representations generated from a sequence of historical wireless observations $\vect{G}_t =
\left[
\vect{g}_{t-\tau+1},
\vect{g}_{t-\tau+2},
\ldots,
\vect{g}_t
\right]$
over a window with length $\tau$.

The first alternative is to generate a time-frequency representation of $\vect{G}_t$ known as a \emph{spectrogram image} \cite{9422831,9674714}.
Here, $\vect{G}_t$ is first divided into a set of overlapping segments using a windowing function (such as a Hann or Hamming window). Then, a Fast Fourier Transform (FFT) is applied to each windowed block, converting the time-domain observations into the frequency domain.
Stacking these frequency domain observations over all segments creates the time-frequency image $\vect{q}_t^{s}$. 

For the second method, we resort to the tools from \gls{tda}, more specifically the \emph{persistence images}.
Here, $\vect{G}_t$ is mapped into an $m$-dimensional point cloud using time-delay embedding \cite{PhysRevLett,6737251}. Next, persistent homology is applied to this point cloud by constructing a simplicial complex (e.g., a Vietoris-Rips complex) that grows with a spatial scale parameter $\epsilon$. This tracks the birth ($\beta$) and death ($\delta$) scales of topological features, such as 1D loops which indicate periodic dynamics, to produce a persistence diagram \cite{edelsbrunner2010computational,892133}. To vectorize this diagram, the coordinates are transformed from birth-death ($\beta$, $\delta$) to birth-persistence ($\beta$, $\delta-\beta$). Finally, a continuous surface is created by placing a 2D Gaussian probability density function over each point \cite{adams2017persistence}, scaled by a weighting function  that typically emphasizes long-lived features, and integrating this surface over a discrete, fixed-size grid to generate the final persistence image (PI) matrix $\vect{q}_t^{p}$.

In this view, the representation that is to be used with the \gls{wjepa} encoder is $\vect{q}_t\in\{\vect{g}_t, \vect{q}_t^{s}, \vect{q}_t^{p} \}$ depending on the use of \gls{csi}, spectrogram, or \gls{pi}.
The exemplary representations are illustrated in Fig. \ref{fig:wireless_embeddings}.
The impact of these three different embeddings on the performance is analyzed under the results.

\begin{figure}
    \centering
    \includegraphics[width=\linewidth]{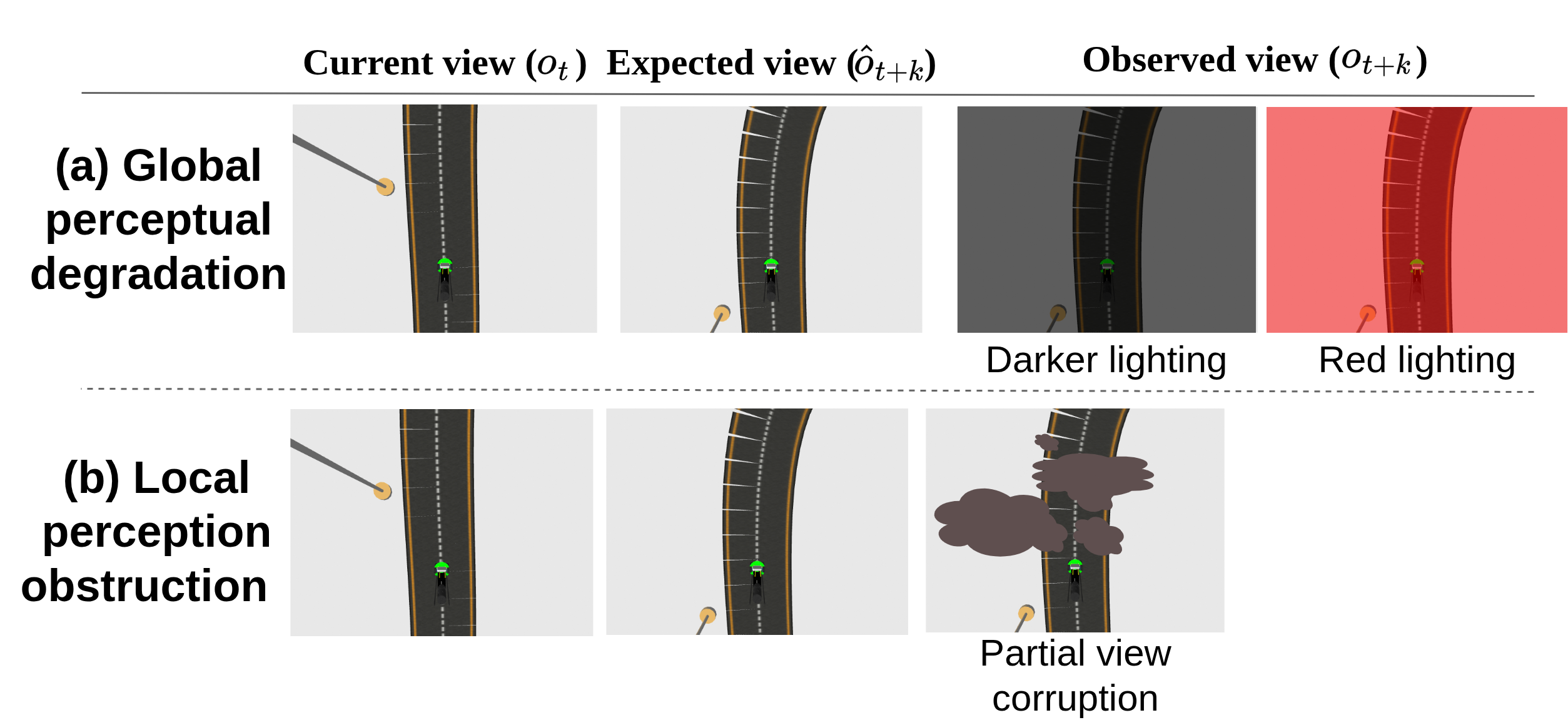}
    \caption{Perception degradation scenarios considered for adaptive resilience: (a) global perceptual degradation caused by illumination variations and (b) local perceptual degradation caused by partial visual occlusions.}
    \label{fig:resilience adaptation}
\end{figure}
\subsection{Adaptive Resilience Mechanism}\label{sec:resilience}
%

Despite of the excessive training during the learning phase discussed earlier, the practical remote robotic systems are likely to encounter environmental changes during the runtime due to undesirable stressors.
Such changes have significance differences from the nominal conditions, which consequently affect the control performance.
Within the scope of this paper, we only focus on two types of runtime changes affecting on the perception: (i) variations on the illumination that are due to global lighting (intensity and color), and (ii) local visual obstruction, such as mud patches, that partially covers the robot's field of view, as depicted in Fig. \ref{fig:resilience adaptation}.
An undesirable level of change can result in inconsistencies between the predictions and perceived observations in the latent space, which is to be utilized within the proposed adaptation mechanism discussed next. 

We first define the discrepancy between the perceived and prediction in the latent space over $K$ future prediction steps considered for inconsistency evaluation as follows:
\begin{equation}
\Delta_t =
\sum\nolimits_{k=1}^{K}
\alpha_k
\left\|
\hat{z}_{t+k \mid t}
-
z_{t+k}
\right\|_2,
\label{eq:latent_inconsistency}
\end{equation}
where $z_{t+k}$ denotes the observed latent representation at time $t+k$, and $\alpha_k$ is the weighting coefficient associated with the $k$-th prediction step.
A perception degradation is identified whenever $\Delta_t \ge \delta$
%
%
where $\delta$ is an empirically selected threshold based on the observed range of latent inconsistencies.

To handle variations on the illumination that are due to global lighting, we first define
%
a finite set of predefined visual transformation parameters
%
$
\Theta = \{\vect{\theta}_1,\vect{\theta}_2,\ldots,\vect{\theta}_M\},
$
%
where each $\vect{\theta}_i$ denotes a predefined \gls{hsv} color transformation tuple. 
Upon noticing a perception degradation,
%
the optimal transformation is selected by minimizing the latent inconsistency as follows:
\begin{equation}
\vect{\theta}^{*}
=
\arg\min_{\theta_i \in \Theta}
\sum_{k=1}^{K}
\alpha_k
\left\|
\hat{z}_{t+k \mid t}
-
\Phi_C
\left(
\mathcal{E}^c
\left(
\mathcal{T}_{\theta_i}(o_t)
\right)
\right)
\right\|_2,
\label{eq:transformation_selection}
\end{equation}
where $\mathcal{T}_{\theta_i}(\cdot)$ denotes the \gls{hsv} color transformation associated with the parameter set $\theta_i$.
The selected transformation is subsequently applied to the incoming observations, thereby restoring latent consistency under varying illumination conditions without modifying the learned \gls{wm}.

For local visual obstruction, 
%
to recover the corrupted visual information, a lightweight \gls{vae} is employed to reconstruct the degraded observation prior to latent encoding,
%
$
\hat{o}_t =
\mathrm{Dec}_{\psi}
\left(
\mathrm{Enc}_{\psi}(o_t)
\right)
$
%
where $\mathrm{Enc}_{\psi}(\cdot)$ and $\mathrm{Dec}_{\psi}(\cdot)$ denote the \gls{vae} encoder and decoder, respectively.
A subset $\mathcal{D}\subset\mathcal{D}_{\mathrm{past}}$ of previously collected observations is utilized for VAE adaptation,
\begin{equation}
r=\frac{|\mathcal{D}|}{|\mathcal{D}_{\mathrm{past}}|},
\qquad
0<r\le1,
\label{eq:replay_dataset}
\end{equation}
where $\mathcal{D}_{\mathrm{past}}$ denotes the repository of previously collected robot observations, $\mathcal{D}$ is the adaptation dataset, and $r$ is the adaptation data ratio.
The \gls{vae} parameters are then optimized as
\begin{equation}
\psi^{*}
=
\arg\min_{\psi}
\mathcal{L}_{\mathrm{VAE}}
\left(
\mathcal{D},
E
\right),
\label{eq:vae_optimization}
\end{equation}
where $E$ denotes the adaptation epochs and $\mathcal{L}_{\mathrm{VAE}}$ represents the standard \gls{vae} reconstruction objective.
By jointly controlling the adaptation ratio $r$ and the optimization epochs $E$, the framework achieves a trade-off between adaptation time and reconstruction performance.
The reconstructed observation is subsequently re-encoded by the \gls{cjepa}, thereby restoring latent consistency under localized visual obstruction.

\subsection{Sionna–Gazebo Integration}
\label{sec:gazebo-sionna}
To evaluate the proposed communication-aware robotic framework under realistic operating conditions, a synchronized robot--wireless simulation environment is developed by integrating the Gazebo robotic simulator with the Sionna  \gls{rt} simulator.
Gazebo provides realistic robot dynamics, perception, and navigation, whereas Sionna \gls{rt} enables physics-based wireless channel modeling through electromagnetic ray tracing.
By combining both simulators, the framework enables synchronized generation of robot observations and wireless channel measurements, allowing the \gls{cjepa} and \gls{wjepa} to jointly predict future robot states and communication conditions within a unified simulation environment.

\begin{figure}
    \centering
    \includegraphics[width=\linewidth]{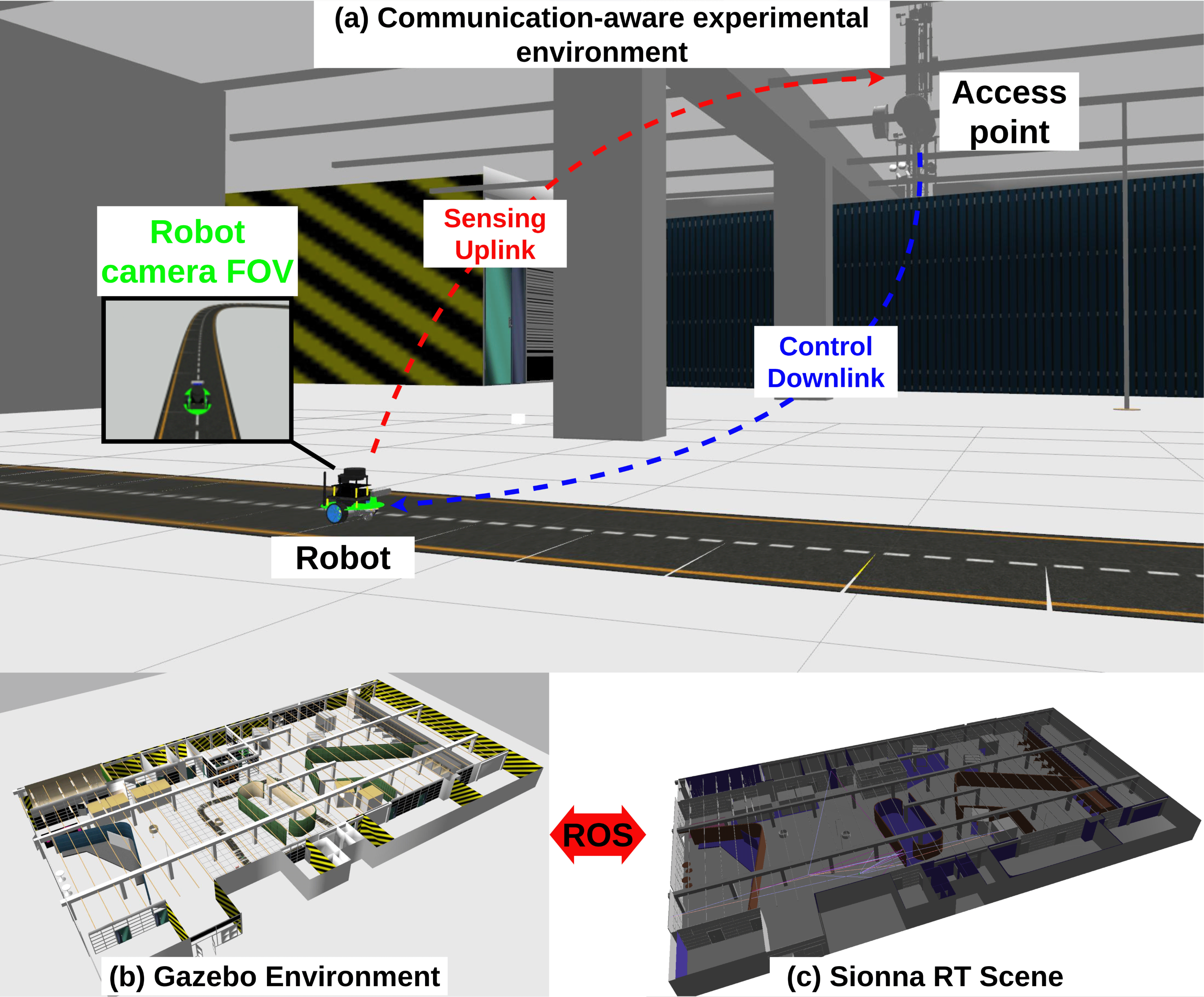}
    \caption{Experimental setup of the proposed communication-aware robotic framework. (a) Overall communication-aware experimental environment with the robot, onboard \gls{rgb} observations, and sensing uplink/control downlink communication. (b) Gazebo environment for robot simulation and visual perception. (c) Corresponding Sionna \gls{rt} scene for physics-based wireless channel generation, synchronized through \gls{ros}.}
    \label{fig:gazebo-sionna}
\end{figure}

Since Gazebo and Sionna \gls{rt} employ different scene representations, the simulation environment cannot be directly shared between the two platforms.
Gazebo represents the environment using \gls{sdf}-based \glspl{wm}, whereas Sionna \gls{rt} requires scenes in the Mitsuba \gls{xml} format for \gls{rt} simulations.
Therefore, as illustrated in Fig. \ref{fig:gazebo-sionna}, the experimental environment is first imported into Blender, where radio material properties are assigned to scene objects before exporting the scene into the Mitsuba \gls{xml} format.
This conversion enables both simulators to represent the same physical environment while supporting realistic wireless propagation modeling \cite{H_P_Gazebo_Sionna_RT_Integration}.

The robotic platform is based on the Waveshare JetBot \gls{ros} \gls{ai} Kit, whose \gls{urdf} model is integrated into the Gazebo environment to provide realistic robot kinematics and onboard \gls{rgb} camera observations \cite{madushanka2026jetbot}.
The \gls{ros} synchronizes robot states between Gazebo and Sionna \gls{rt}, enabling the robot pose to be continuously supplied to the ray-tracing engine for \gls{csi} generation while simultaneously providing visual observations to the \gls{cjepa}.
Consequently, the integrated simulation environment supports synchronized evaluation of robot perception, wireless communication, and communication-aware remote robotic control under realistic operating conditions.

\subsection{Experimental Setup}

%
Within the communication-aware synchronized robot–wireless simulation environment, Gazebo and Sionna \gls{rt} operate synchronously through \gls{ros}, enabling simultaneous generation of robot observations and wireless channel measurements.
At each simulation step, the robot pose and robot velocity are shared between both simulators, allowing the \gls{rgb} observation acquired by the onboard camera and the corresponding \gls{csi} generated by Sionna \gls{rt} to represent the same robot state.
These synchronized observations are subsequently processed by the \gls{cjepa} and \gls{wjepa} for joint prediction and communication-aware decision making.

Figure \ref{fig:gazebo-sionna} illustrates the overall experimental setup employed throughout the evaluation. 
As shown in Fig. \ref{fig:gazebo-sionna}(a), a mobile robot equipped with an onboard \gls{rgb} camera performs autonomous navigation within a closed-loop road environment while maintaining bidirectional wireless communication with a fixed base station through the sensing uplink and control downlink.
Figures \ref{fig:gazebo-sionna}(b) and \ref{fig:gazebo-sionna}(c) present the corresponding Gazebo and Sionna \gls{rt} environments, respectively, where the same robot state is synchronized between both simulators to jointly generate visual observations and wireless channel measurements.

The implementation is developed using \gls{ros} Noetic, Gazebo, Sionna \gls{rt}, Mitsuba, Blender, Python, and PyTorch. The detailed hardware specifications, software versions, wireless communication parameters, and learning hyperparameters used throughout the experiments are provided in Appendix \ref{apndx:parameters}.

\subsection{Transfer Learning and Domain Adaptation}

Training the \gls{cjepa} directly in the Gazebo environment from random initialization is computationally expensive because each interaction requires realistic physics simulation, robot dynamics, \gls{ros} communication, and onboard camera rendering.
Consequently, training the navigation model entirely in Gazebo requires several days of continuous interaction before convergence.
To improve training efficiency, the \gls{cjepa} is first pretrained in the lightweight Gym CarRacing environment \cite{towers2026gymnasium}, where navigation experiences can be generated rapidly.
The pretrained model is then transferred to the Gazebo environment and fine-tuned using Gazebo interaction data rather than training a new model from scratch.
After convergence, the adapted \gls{cjepa} is frozen and subsequently used to generate latent control representations for training the \gls{wjepa}.

A major challenge during this transfer process is the mismatch between the observation spaces of the two environments.
The pretrained \gls{cjepa} learns visual representations from the Gym CarRacing observation space, whereas the Gazebo robot acquires realistic \gls{rgb} images from its onboard camera.
%
%
Therefore, instead of modifying the network architecture, the Gazebo robot observation is transformed into a Gym-style observation before being processed by the \gls{cjepa},
$
o_t =
\mathcal{T}_{\theta}
\left(
o_t^{G}
\right)
$,
where $o_t^G$ denotes the raw \gls{rgb} observation captured by the Gazebo robot, and $\mathcal{T}_{\theta}(\cdot)$ is the observation transformation parameterized by the predefined image-processing parameters $\theta$, including the \gls{hsv} color thresholds and other transformation settings used to generate the Gym-style observations.
The transformation extracts the drivable road region, suppresses irrelevant visual information, and generates a normalized Gym-style observation that is compatible with the pretrained \gls{cjepa}. 
%

Unlike the Gym CarRacing environment, the Gazebo simulator does not provide an intrinsic \gls{rl} reward. 
Therefore, a progress-based reward function is designed to encourage continuous road following while discouraging inefficient or unsafe robot behaviors.
The reference trajectory is partitioned into uniformly distributed virtual crossing markers, and the robot receives a positive reward whenever it crosses a previously unvisited marker.
Since the Gazebo control loop operates at approximately $35$\,ms per simulation step, a small step penalty is applied at every interaction to encourage efficient navigation and reduce unnecessary exploration.
The reward function is defined as
\begin{equation}
r_t
=
r_{\mathrm{step}}
+
\frac{R_{\mathrm{lap}}}{N_{\mathrm{cross}}}
n_t
-
R_{\mathrm{fail}}
\Pi_{\mathrm{fail}},
\label{eq:reward_function}
\end{equation}
where 
$r_{\mathrm{step}}$ denotes the per-step penalty, 
$R_{\mathrm{lap}}$ is the reward assigned for completing one lap, 
$N_{\mathrm{cross}}$ represents the total number of virtual crossing markers, 
$n_t$ is the number of newly crossed markers at time $t$, and
$\Pi_{\mathrm{fail}}$ is a binary indicator corresponding to episode termination.
An episode is terminated when the robot deviates from the road, remains stationary for an extended period due to excessive braking, overturns following a collision or instability, or when the cumulative reward falls below a predefined threshold. 
Upon termination, an additional failure penalty is assigned to discourage unsafe and undesirable navigation behaviors.
This reward formulation preserves the progress-driven objective of Gym CarRacing while adapting it to autonomous navigation in the Gazebo environment.

The pretrained \gls{cjepa} predicts navigation decisions using the discrete action space $\mathcal{A}_d=\{0,1,2,3,4\}$ adopted by the Gym CarRacing environment, which includes the actions coast, steer right, steer left, accelerate, and brake. 
However, the Gazebo JetBot is controlled through continuous linear $v_t$ and angular velocity $\omega_t$ commands. 
To bridge this interface mismatch, a discrete-to-continuous action mapping $(v_t,\omega_t) = \mathcal{M}(a_t)$ is introduced,
%
%
%
and the linear and angular velocities transmitted through the \gls{ros} \texttt{cmd\_vel} topic.
To ensure physically feasible motion, the generated commands are further constrained using smooth acceleration, braking, and steering-rate limits before being executed by the robot.
This action conversion enables the policy learned in the discrete Gym environment to be directly deployed in the continuous-control Gazebo environment.




\section{Performance Evaluation}
\label{sec:results}

The proposed 
framework is evaluated
and compared with two baseline methods
through a comprehensive set of experiments within the synchronized Gazebo–Sionna simulation environment described in Section \ref{sec:sys_mod}.
The evaluation investigates the ability of the proposed \gls{cjepa} and \gls{wjepa} to maintain reliable remote robotic control under limited wireless connectivity while reducing communication resource utilization.
To this end, we first introduce the experimental evaluation scenarios and baseline methods used for comparison.
Furthermore, the experiments analyze
diverse wireless channel variations and visual perception degradations to assess the robustness, adaptive resilience, and predictive wireless world modeling capability.
%
Towards this we used an array of evaluation metrics summarized in Table \ref{tab:evaluation_metrics}.

\begin{table}[!t]
\caption{Evaluation metrics used to assess the frameworks.}
\label{tab:evaluation_metrics}
\centering
\begin{tabular}{l p{.575\linewidth}}
\toprule
\textbf{Metric} & \textbf{Definition}  \\
\midrule
$T_c=t_f^{\mathrm{sim}}-t_0^{\mathrm{sim}}$ 
&
Task (navigation) completion time \\

$N_{\mathrm{comm}}=\sum_{t=1}^{T}\rho_t$ 
&
Communication (uplink) rounds\\

$E_{\mathrm{tot}}=\sum_{i=1}^{N_{\mathrm{comm}}}P_i\Delta t$ 
&
Total transmission (uplink) energy\\

$R=\sum_{t=1}^{T}r_t$ 
&
Accumulated task reward \\
\bottomrule
\end{tabular}
\end{table}

\subsection{Experimental Evaluation Scenarios}

\newcommand{\expOrginal}{\textsc{E1}}
\newcommand{\expAddScat}{\textsc{E2}}
\newcommand{\expChngAP}{\textsc{E3}}
\newcommand{\expChngFreq}{\textsc{E4}}
\newcommand{\expModCSI}{\textsc{E5}}
\newcommand{\expPredict}{\textsc{E6}}
\newcommand{\expChngLit}{\textsc{E7}}
\newcommand{\expChangFoV}{\textsc{E8}}

\begin{figure}
    \centering
    \includegraphics[width=\linewidth]{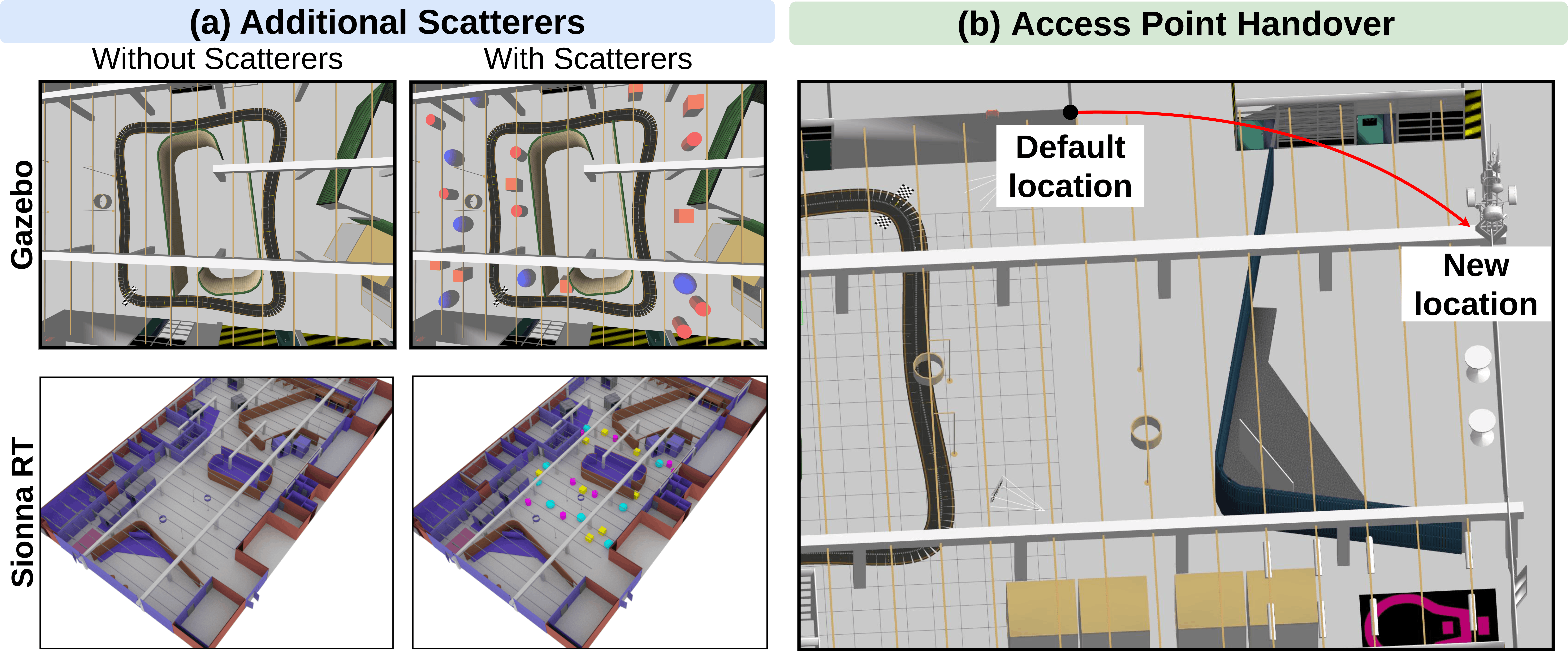}
    \caption{Wireless evaluation scenarios employed to assess the robustness of the proposed communication-aware framework: (a) Additional scatterers are introduced into both the Gazebo and Sionna \gls{rt} scenes to generate diverse multipath propagation conditions; and (b) The access-point handover is enforced to modify the geometry of the wireless environment.}
    \label{fig:evaluation_scenarios}
\end{figure}
To systematically evaluate the proposed communication-aware robotic framework, a comprehensive set of simulation scenarios, referred to as \expOrginal-\expChangFoV, is designed within the synchronized Gazebo–Sionna simulation environment.
Throughout all experiments, the robot follows the same predefined closed-loop trajectory, while the wireless communication environment or visual perception conditions are selectively modified according to the evaluation objective.
\expOrginal{} is the original experiment where training and testing settings are statistically identical.
Towards evaluating the robustness against changing wireless channel propagation characteristics, geometry, and line-of-sight conditions, 
we consider 
introducing additional scatterers in \expAddScat{} (see Fig. \ref{fig:evaluation_scenarios}a),
simulating access point handover in \expChngAP{} (see Fig. \ref{fig:evaluation_scenarios}b),
and
changing carrier frequency of $2.14$\,GHz used in training to $1.9$\,GHz and $2.6$\,GHz for the testing under \expChngFreq.
The advantages of using structured \gls{rf} representations, spectrogram- and \gls{pi}-based representations, over the use of the conventional raw \gls{csi} with \gls{wjepa} is evaluated in \expModCSI.
With \expPredict, the predictive capabilities under consecutive connectivity failures are analyzed.
Finally, the adaptability to different lighting conditions and visual obstructions discussed in Sec. \ref{sec:resilience} are evaluated under \expChngLit{} and \expChangFoV, respectively.

\subsection{Baseline Methods}
To evaluate the effectiveness of the proposed communication-aware robotic framework, two representative navigation baselines are considered: a \gls{dqn} and a \gls{pid} controller.
Both baselines are implemented within the same synchronized Gazebo–Sionna simulation environment and evaluated under identical experimental conditions.
To ensure a fair comparison, the \gls{dqn} baseline follows the same transfer learning strategy adopted by the proposed \gls{cjepa}. 
%
%
This enables the \gls{dqn} to operate under the same observation space and training protocol as the proposed framework.

The \gls{pid} controller serves as a conventional feedback-control navigation baseline. 
It directly follows the predefined reference trajectory using the robot's onboard camera observations to generate continuous steering commands without predictive world modeling or communication-aware decision-making. 
This baseline provides a reference for evaluating the benefits of the proposed predictive and communication-aware control framework.

For both baseline methods, the robot continuously transmits the onboard \gls{rgb} observations to the remote controller through the sensing uplink, while the corresponding control commands are returned through the control downlink at every control interval. 
Unlike the proposed framework, neither baseline employs predictive wireless modeling, transmission scheduling, or wireless power prediction. 
Consequently, they provide communication-unaware references for evaluating the communication efficiency achieved by the proposed framework.

\subsection{Overall Task Performance \& Robustness}

\begin{table}[!t]
\caption{The comparison of the overall performance under \expOrginal.}
\label{tab:overall_performance}
\centering
\begin{tabular}{llccc}
\toprule
\multicolumn{2}{l}{\textbf{Metric}}
& \textbf{Proposed} 
& \textbf{DQN}
& \textbf{PID} \\
\midrule
\multicolumn{2}{l}{Completion time (s)}
& $57.1 \pm 1.0$ & $58.6 \pm 1.6$ & $77.3 \pm 0.3$ \\
\multicolumn{2}{l}{Total reward}
& $775.4 \pm 4.4$ & $770.8 \pm 3.5$ & $716.1 \pm 2.0$ \\
\multirow{2}{20pt}{Latency (ms)}& Uplink
& $0.48 \pm 0.02$ & $0.86 \pm 0.00$ & $2.63 \pm 0.05$  \\
& Downlink
& $0.44 \pm 0.00$ & $0.44 \pm 0.00$ & $1.31 \pm 0.02$  \\
\multicolumn{2}{l}{Total energy (J)}
& $4.74 \pm 0.16$ & $9.21 \pm 0.32$ & $11.19 \pm 0.15$  \\
\bottomrule
\end{tabular}
\end{table}

Table~\ref{tab:overall_performance} summarizes the overall navigation and communication performance of the proposed framework in comparison with the \gls{dqn} and \gls{pid} baselines within \expOrginal{} setting. 
The proposed framework achieves the shortest task completion time of $57.1$\,s while obtaining the highest cumulative average reward of $775.4$, demonstrating efficient and stable navigation. 
Although \gls{dqn} exhibits comparable navigation performance, the proposed method further reduces the completion time and improves the cumulative reward. 
In contrast, the \gls{pid} controller requires the longest completion time ($77.3$\,s on average) and achieves the lowest average reward of $716.1$, indicating inferior navigation performance. 
Furthermore, the proposed framework attains the lowest communication latency, particularly in the uplink direction with $0.48$\,ms, while maintaining a comparable downlink latency of $0.44$\,ms. 
This communication efficiency is achieved through predictive and selective uplink transmissions, which also result in the lowest cumulative uplink transmission energy of $4.74$\,J, compared with $9.21$\,J for \gls{dqn} and $11.19$\,J for \gls{pid}. 
Overall, the results demonstrate that the proposed communication-aware predictive control framework effectively balances navigation performance, communication latency, and energy efficiency.

\begin{figure}
    \centering

    \begin{subfigure}{0.158\textwidth}   
        \centering
        \includegraphics[width=\linewidth]{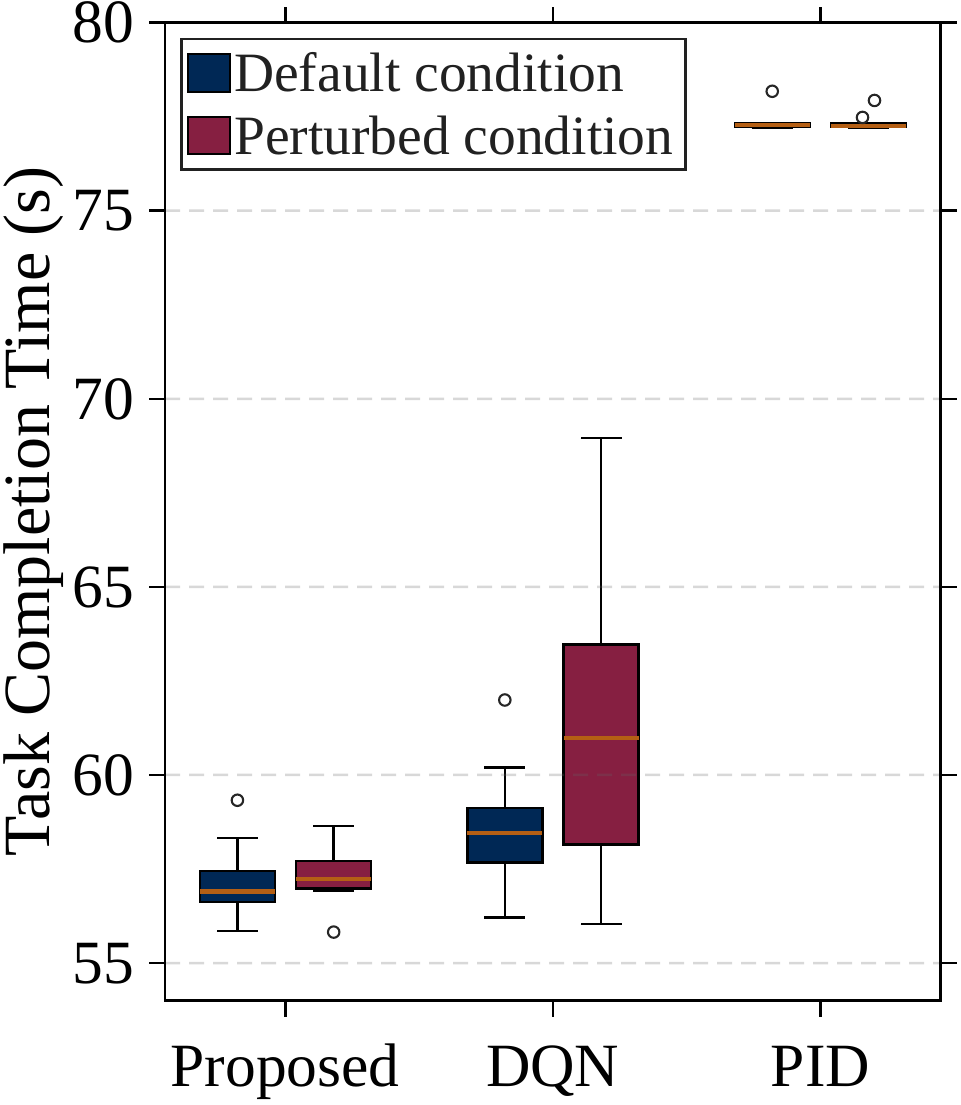}
        \caption{1.9 GHz}
        \label{fig:time_19ghz}
    \end{subfigure}
    \hfill
    \begin{subfigure}{0.158\textwidth}
        \centering
        \includegraphics[width=\linewidth]{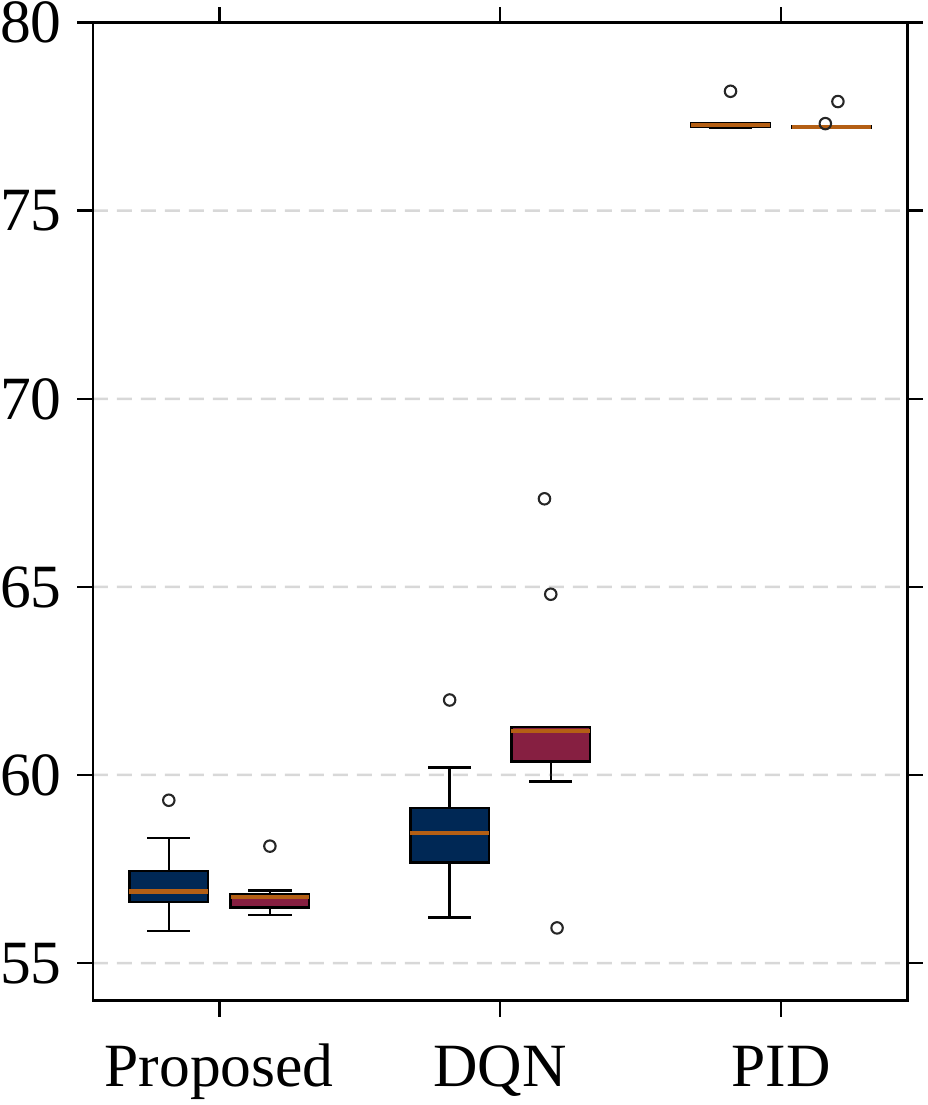}
        \caption{2.6 GHz}
        \label{fig:time_26ghz}
    \end{subfigure}
    \hfill
    \begin{subfigure}{0.158\textwidth}
        \centering
        \includegraphics[width=\linewidth]{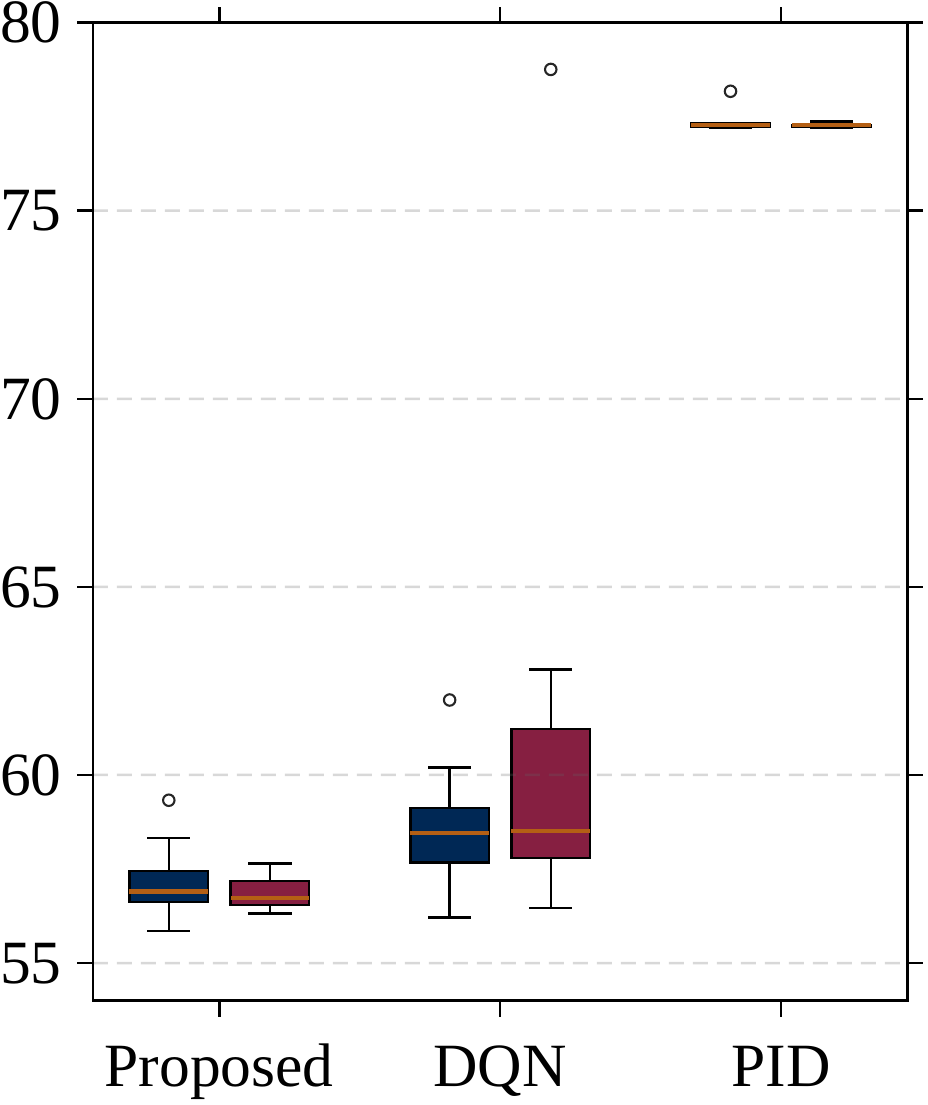}
        \caption{Added scatterers}
        \label{fig:time_scatterers}
    \end{subfigure}

    \caption{Task completion time under wireless channel perturbations:
    (a) carrier frequency shifted from 2.14 GHz to 1.9 GHz,
    (b) carrier frequency shifted from 2.14 GHz to 2.6 GHz, and
    (c) additional scatterers introduced into the propagation environment.}
    \label{fig:robust_task_com}
\end{figure}

\begin{figure}
    \centering

    \begin{subfigure}{0.158\textwidth}   
        \centering
        \includegraphics[width=\linewidth]{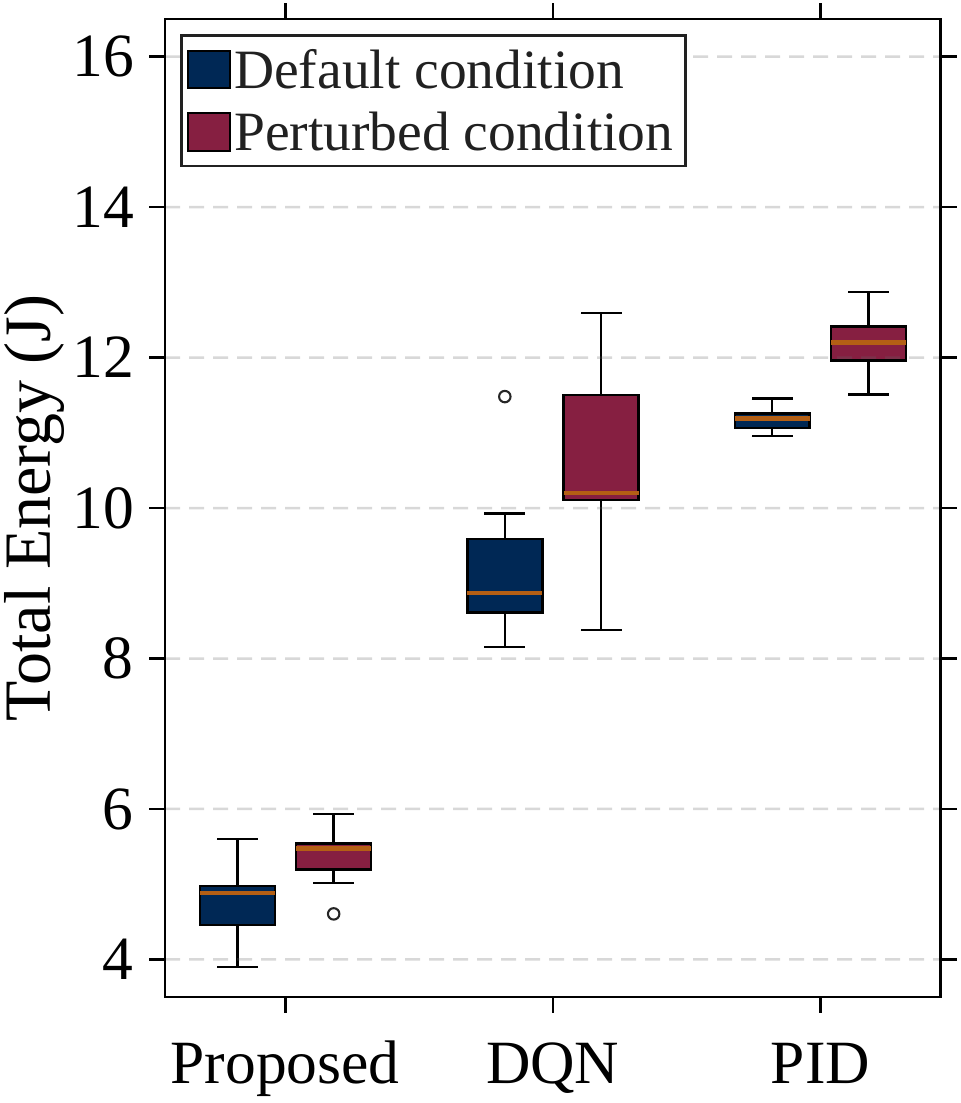}
        \caption{1.9 GHz}
        \label{fig:energy_19ghz}
    \end{subfigure}
    \hfill
    \begin{subfigure}{0.158\textwidth}
        \centering
        \includegraphics[width=\linewidth]{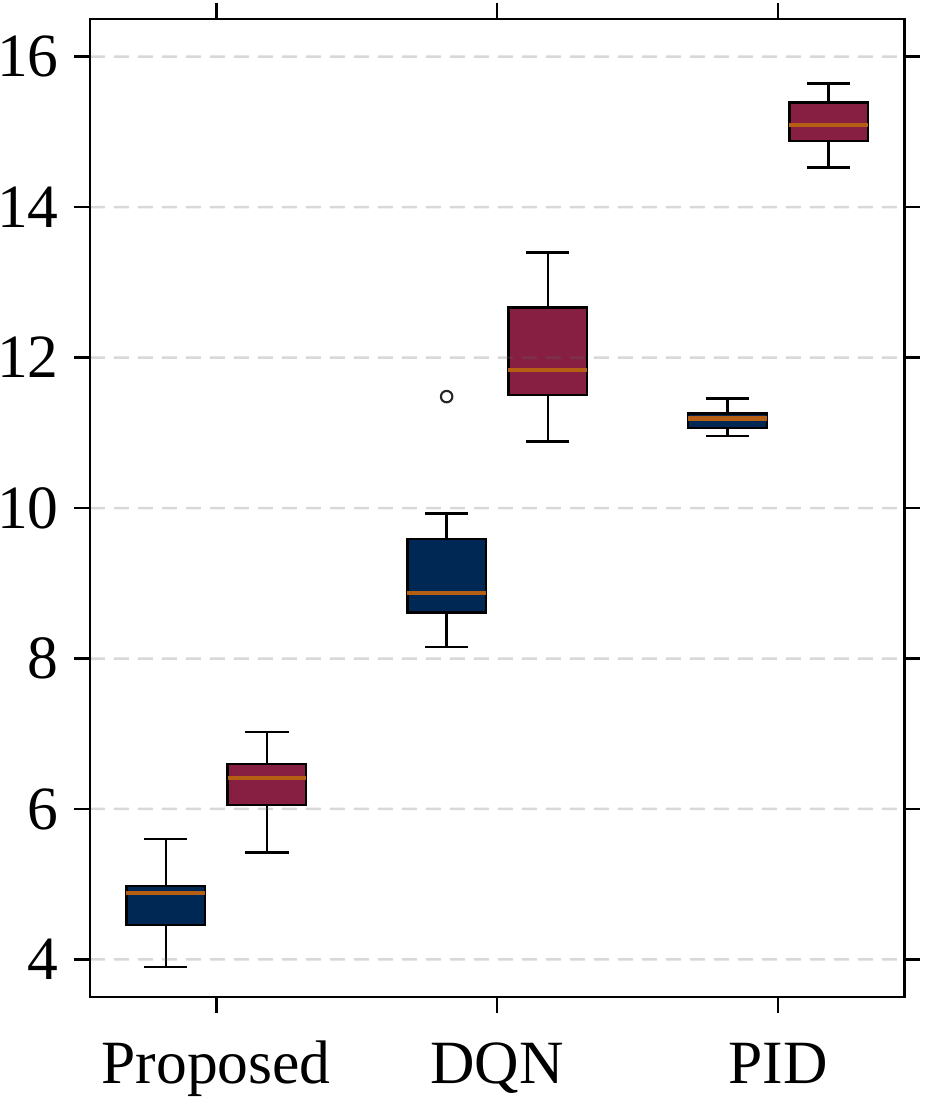}
        \caption{2.6 GHz}
        \label{fig:energy_26ghz}
    \end{subfigure}
    \hfill
    \begin{subfigure}{0.158\textwidth}
        \centering
        \includegraphics[width=\linewidth]{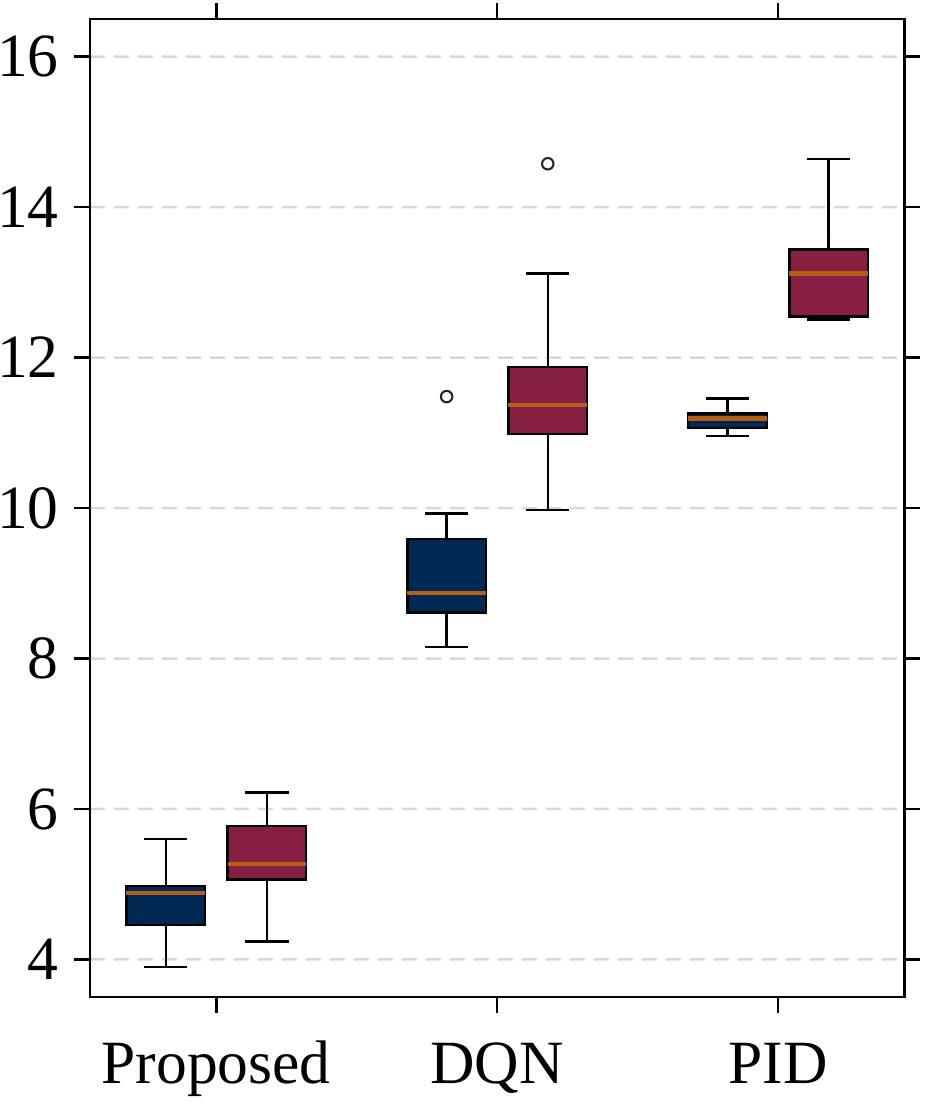}
        \caption{Added scatterers}
        \label{fig:energy_scatterers}
    \end{subfigure}

    \caption{Total uplink transmission energy under wireless channel perturbations:
    (a) carrier frequency shifted from 2.14 GHz to 1.9 GHz,
    (b) carrier frequency shifted from 2.14 GHz to 2.6 GHz, and
    (c) additional scatterers introduced into the propagation environment.}
    \label{fig:robust_power_consumption}
\end{figure}

The robustness of the proposed communication-aware control framework is evaluated with \expChngFreq, \expAddScat, and \expChngAP.
Figs. \ref{fig:robust_task_com} and \ref{fig:robust_power_consumption} compare the task completion time and the total uplink transmission energy consumption, respectively, of the proposed method and \gls{dqn} and \gls{pid} baselines under two carried frequency changes as per \expChngFreq{} and additional scatterers as per \expAddScat.
The proposed framework exhibits only minor variations in task completion time across all perturbation scenarios, demonstrating strong generalization to unseen wireless environments.
In contrast, the \gls{dqn} baseline shows larger performance degradation, whereas the \gls{pid} controller remains relatively unaffected since its control policy does not rely on learned communication-aware representations.
Furthermore, the proposed framework consistently achieves the lowest uplink transmission energy by selectively scheduling transmissions according to the predicted communication requirements, while the \gls{dqn} and \gls{pid} baselines incur substantially higher energy consumption.
Overall, these results demonstrate that the proposed communication-aware predictive control framework preserves navigation performance while maintaining superior communication energy efficiency under varying wireless propagation conditions.


To further evaluate robustness, the serving access point is switched to emulate a handover event in \expChngAP, producing a substantial change in wireless coverage.
%
We report that the proposed framework successfully completes ($100\%$) the navigation task despite the access-point handover, whereas both the \gls{dqn} and \gls{pid} baselines only manage to complete $56.48\%$ and $58.99\%$ of the lap, respectively. 
%
This demonstrates the robustness of the proposed framework under significant changes in the wireless network topology.

\subsection{Analysis of The Structured \gls{rf} Representations}

While the proposed \gls{wjepa} design relies on the raw \gls{csi} measurements as the input, here, we analyze the impact of using two structured \gls{rf} representations, spectrograms and \glspl{pi} discussed in Sec. \ref{sec:wireless_embeddings}, as alternative inputs within \expModCSI.
%

\begin{figure}
    \centering
    \includegraphics[width=\linewidth]{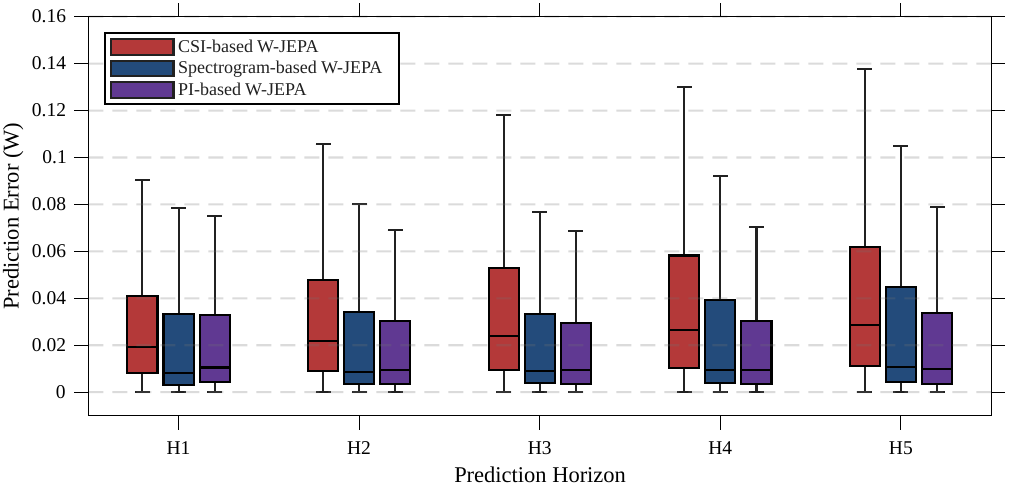}
    \caption{Prediction error comparison between raw \gls{csi}-based and structured representation-based \gls{wjepa} models across multiple prediction horizons.}
    \label{fig:all_methods}
\end{figure}

As shown in Fig. \ref{fig:all_methods} both structured \gls{rf} representations consistently reduce the prediction error compared with the raw \gls{csi}-based \gls{wjepa} across all prediction horizons ($H$).
The improvement becomes more evident as $H$ increases, indicating that the structured \gls{rf} representations provide richer temporal information for learning future wireless channel dynamics.
Among the proposed representations, the \gls{pi}-based \gls{wjepa} achieves the lowest prediction error distribution, while the spectrogram-based \gls{wjepa} also demonstrates a clear improvement over the raw \gls{csi}-based model.

\begin{figure}
    \centering
    \begin{subfigure}{0.241\textwidth}   
        \centering
        \includegraphics[width=\linewidth]{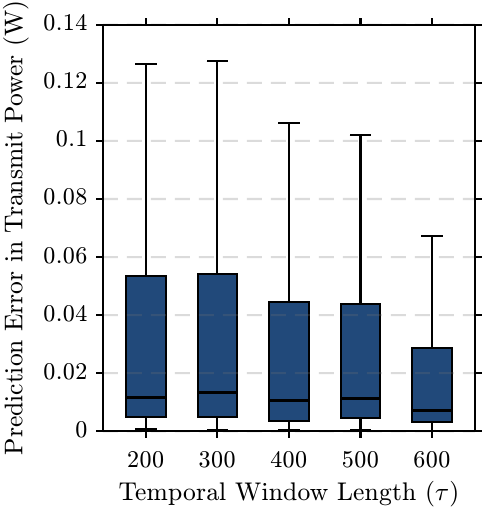}
        \caption{Spectrogram-based \gls{wjepa}}
        \label{fig:Spectrogram_images}
    \end{subfigure}
    \hfill
    \begin{subfigure}{0.241\textwidth}
        \centering
        \includegraphics[width=\linewidth]{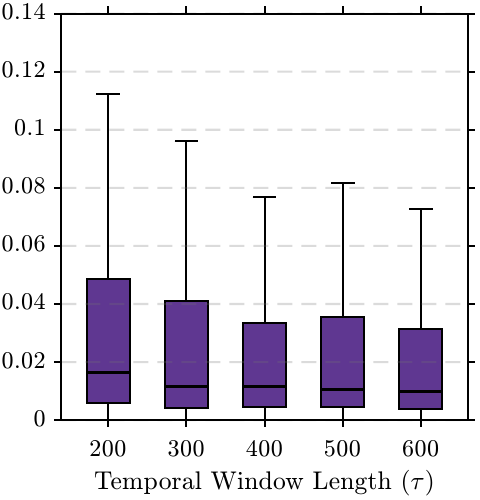}
        \caption{\Gls{pi}-based \gls{wjepa}}
        \label{fig:PI_images}
    \end{subfigure}
    \caption{Impact of temporal window length ($\tau$) on uplink transmit-power prediction error for
    (a) spectrogram-based \gls{wjepa} and 
    (b) \gls{pi}-based \gls{wjepa}.
    }
    \label{fig:temporal_window}
\end{figure}

The influence of the temporal window length $\tau$ is evaluated in Fig. \ref{fig:temporal_window} for both structured \gls{rf} representations. 
Increasing the temporal window provides additional historical channel information, leading to improved prediction performance. 
%
However, for both representations, longer temporal windows generally result in lower prediction errors and reduced variability, with the lowest error distributions observed at the larger window lengths.

\newcommand{\genTime}{T_{\textbf{G}}}

\begin{table}[!t]
\caption{Impacts of representation lengths on power prediction errors ($\mathcal{L}_{P}$) and representation generation times ($\genTime$).}
\label{tab:structured_RF}
\centering
\begin{tabular}{ccc p{0pt} ccc}
\multicolumn{3}{c}{\textbf{Spectrogram-based \gls{wjepa}}} && \multicolumn{3}{c}{\textbf{PI-based \gls{wjepa}}}\\
\cmidrule{1-3} \cmidrule{5-7}
$L$ & $\mathcal{L}_{P}$ (W) & $\genTime$ (ms) && $m$ & $\mathcal{L}_{P}$ (W) & $\genTime$ (ms)\\
\cmidrule{1-3} \cmidrule{5-7}
64 & $0.022 \pm 0.026$ & $1.74$ && 5 & $0.020 \pm 0.021$ & $98.84$  \\
96 & $0.014 \pm 0.018$ & $1.71$ && 10 & $0.019 \pm 0.020$ & $99.9$   \\
128 & $0.022 \pm 0.025$ & $1.66$ && 15 & $0.025 \pm 0.033$ & $102.03$   \\
160 & $0.024 \pm 0.027$ & $1.65$ && 20 & $0.014 \pm 0.013$ & $104.46$   \\
200 & $0.024 \pm 0.028$ & $1.63$ && 25 & $0.028 \pm 0.030$ & $103.45$   \\
\cmidrule{1-3} \cmidrule{5-7}
\end{tabular}
\end{table}

Table~\ref{tab:structured_RF} summarizes the impact of the representation-generation parameters and the corresponding computational complexity of the proposed structured \gls{rf} representations.
For the spectrogram-based representation, the \gls{stft} segment length ($L$) influences both the uplink power prediction error ($\mathcal{L}_{P}$) and the representation generation time ($\genTime$), while the time-delay embedding dimension ($m$) provides a similar tradeoff for the \gls{pi} representation.
The selected parameter settings achieve an effective balance between prediction accuracy and computational complexity, while both representations remain computationally efficient for online operation.
Overall, these results demonstrate that the proposed structured \gls{rf} representations provide an effective foundation for predictive wireless-world modeling, significantly improving the predictive capability of the \gls{wjepa} without compromising real-time applicability.

These results reveal a tradeoff between representation richness and computational complexity. Raw \gls{csi} avoids the additional preprocessing required by the structured \gls{rf} representations, but exhibits higher prediction errors in our experiments. 
The spectrogram represents the temporal channel history in the time-frequency domain and achieves improved prediction accuracy with a relatively small representation-generation time. 
The \gls{pi} provides a topology-based representation of the temporal channel evolution and achieves the lowest prediction errors, although at a substantially higher representation-generation cost, as shown in Table~\ref{tab:structured_RF}. 
Consequently, raw \gls{csi} is attractive when representation-processing overhead is the primary constraint, the spectrogram provides a favorable balance between prediction accuracy and computational complexity, whereas the \gls{pi} is preferable when improved predictive performance justifies the additional computational cost.

\subsection{Communication Efficiency Through Predictive Latent Embeddings}

To evaluate the communication efficiency of the proposed framework, we compare the predictive \gls{jepa} latent embeddings with the image token representations employed by the \gls{vit}-based baseline within \expPredict. 
The evaluation therein considers both the transmitted token size and the robustness to consecutive uplink communication losses.
\begin{figure}
    \centering
    \includegraphics[width=\linewidth]{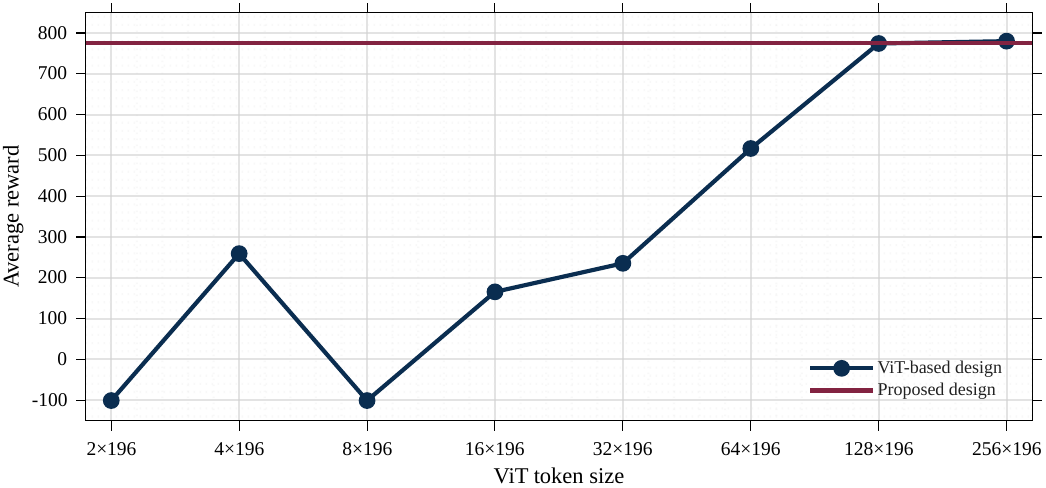}
    \caption{Impact of transmitted token size on reward for the proposed framework and the \gls{vit}-based baseline.}
    \label{fig:token_size}
\end{figure}

As shown in Fig. \ref{fig:token_size}, the proposed framework employs a fixed latent embedding of 400 dimensions, whereas the x-axis is extended to represent the equivalent token sizes of the \gls{vit}-based baseline for comparison.
It can be noted that the \gls{vit} baseline requires at least $128 \times 196$ tokens to achieve comparable performance.
This demonstrates that the predictive \gls{jepa} embeddings provide a substantially more compact representation, significantly reducing the communication payload by about $62$-folds in terms of the size without sacrificing control performance.

\begin{figure}
    \centering
    \includegraphics[width=\linewidth]{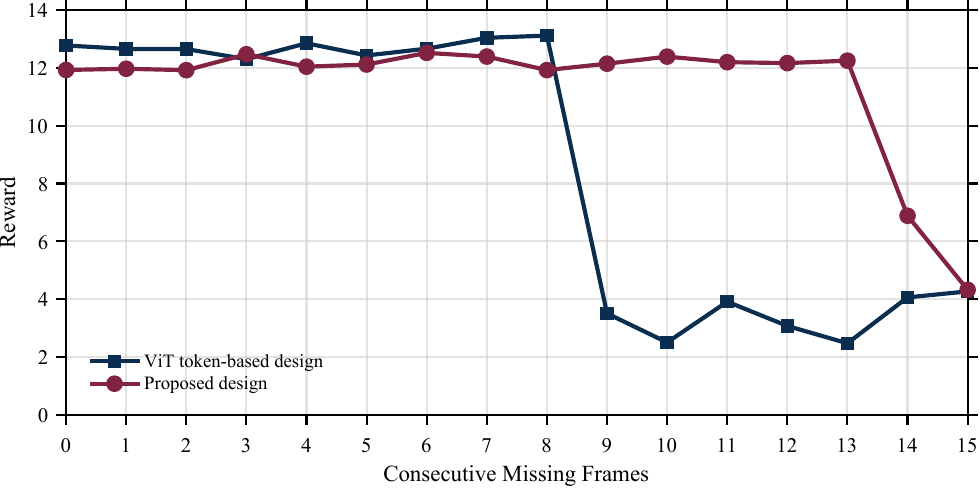}
    \caption{Impact of consecutive uplink communication losses on the achieved reward for the proposed framework and the \gls{vit}-based baseline.}
    \label{fig:mising_frame}
\end{figure}

Fig. \ref{fig:mising_frame} compares the impact of consecutive uplink communication losses on the achieved reward.
The \gls{vit}-based baseline maintains reliable performance for approximately $8$ consecutive missing transmissions, after which the reward rapidly deteriorates.
In contrast, the proposed framework sustains stable performance for approximately $13$ consecutive missing transmissions by exploiting predictive latent dynamics learned through the \gls{jepa} \gls{wm}.
These results demonstrate that the proposed predictive latent embeddings not only reduce communication overhead but also improve resilience to intermittent wireless connectivity.

\subsection{Adaptive Resilience to Visual Disturbances}

%

\begin{figure}
    \centering
    \includegraphics[width=\linewidth]{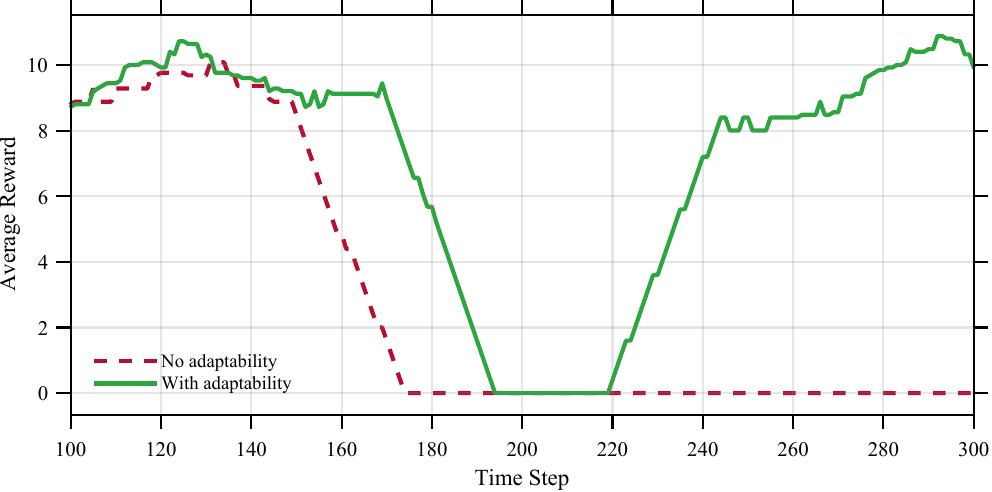}
    \caption{Impact of adaptive latent representation alignment on reward under global perceptual degradation caused by illumination changes.}
    \label{fig:resilience_1}
\end{figure}

The adaptive resilience of the proposed framework is evaluated under \expChngLit{} and \expChangFoV.
The average reward with and without adaptation under \expChngLit{} is compared in Fig. \ref{fig:resilience_1}.
Therein, changes in the illumination conditions alter the robot's field of view, causing the controller without adaptability to fail due to the resulting perceptual inconsistency. 
In contrast, the proposed adaptive resilience mechanism successfully detects the latent inconsistency introduced by the illumination change and restores perceptual consistency through online adaptation as discussed in Sec. \ref{sec:resilience}. 
Consequently, the robot is in capable of continuing the navigation task with a slight interruption.

\begin{figure}
    \centering
    \includegraphics[width=\linewidth]{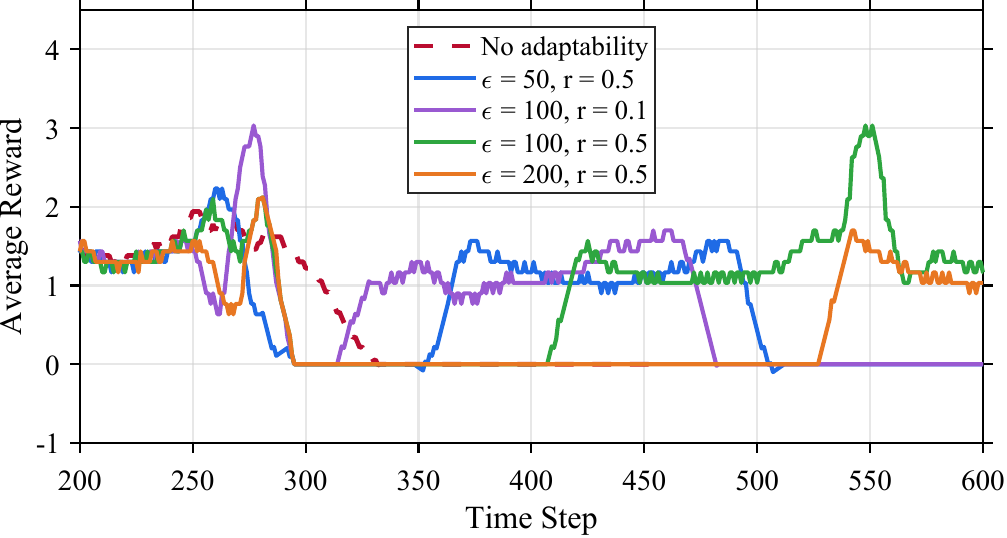}
    \caption{Impact of historical data ratio ($r$) and adaptation epochs ($\epsilon$) on recovery from local visual obstruction.}
    \label{fig:resilience_2}
\end{figure}

The recovery capability under local visual obstruction as per \expChangFoV{} is evaluated in Fig. \ref{fig:resilience_2}.
Therein, we analyze the impact of the fraction of historical observations stored on the server used for online adaptation ($r$) and the number of epochs used for adaptation ($\epsilon$) on the reward. 
Note that higher the $r$ and/or $\epsilon$, the adaptation time increases due to the need for increased computation.
From Fig. \ref{fig:resilience_2}, it can be seen that the faster recovery (with lower $r$ and $\epsilon$) is not always sustainable.
This yields that a significant computation with either one or both of $r$ and $\epsilon$ is needed to ensure that the performance is maintained for a longer duration, i.e., an interesting tradeoff between recovery and sustainability with the proposed adaptation mechanism.

%
%

Overall, the proposed adaptive resilience mechanism enables reliable recovery from both global perceptual degradation and local visual obstruction without retraining the complete control policy. 
By restoring latent consistency online, the framework maintains stable robot navigation and improves operational continuity under challenging visual conditions.

\section{Conclusion and Future Work}
\label{sec:conclusion}

This paper presented a resilient communication-aware remote robotic control framework based on coupled \gls{cjepa} and \gls{wjepa} predictive \glspl{wm}. By jointly modeling robot dynamics and wireless channel evolution within a shared latent representation space, the proposed framework enables predictive communication scheduling that reduces unnecessary uplink transmissions while maintaining reliable navigation performance. Furthermore, structured \gls{rf} representations improve wireless prediction accuracy, while the adaptive resilience mechanism allows the framework to recover from wireless perturbations and visual perception degradations without retraining the complete control policy. Experimental evaluation in a realistic Gazebo--\gls{ros}--Sionna simulation environment demonstrated that the proposed approach consistently outperforms conventional \gls{dqn}, \gls{pid}, and raw \gls{csi}-based methods in communication efficiency, robustness, resilience, and predictive wireless modeling.

The proposed framework also highlights the practical trade-off between communication efficiency and computational complexity. Although predictive world modeling, structured \gls{rf} representation generation, and online adaptation introduce additional computation, the measured execution times demonstrate that the framework remains suitable for real-time operation on a single \gls{gpu}. Moreover, the integrated simulation framework provides a practical platform for evaluating communication-aware robotic systems under realistic wireless propagation conditions before physical deployment.

The current study is limited to a single robot operating in simulation over a single wireless link and predefined navigation scenarios. Future work will extend the framework to multi-robot systems, heterogeneous wireless networks, and more diverse operating environments while investigating additional structured \gls{rf} representations and more efficient online adaptation strategies. Finally, the proposed framework will be validated on physical robotic platforms operating over real wireless networks and extended toward a communication-aware digital twin through continuous synchronization between the physical and virtual environments.

\section*{Code and Data Availability}

The source code of the proposed communication-aware remote robotic control framework is publicly available at \url{https://github.com/ICONgroupCWC/DreamerV2-Meets-Gazebo}.
A demonstration video illustrating the proposed framework and the experimental evaluation presented in this paper is available at
\url{https://youtu.be/hw_bdS3P6Oc}.


\bibliographystyle{IEEEtran}
\bibliography{references}

\appendices
\section{Implementation Details}\label{apndx:parameters}

\begin{table}[!ht]
\caption{Hardware Configuration}
\label{tab:hardware}
\centering
\begin{tabular}{ll}
\hline
\textbf{Component} & \textbf{Specification} \\
\hline
CPU & Intel 11th Gen Core i5-11600K @ 3.90 GHz \\
&(6 cores, 12 threads, up to 4.90 GHz) \\
GPU & NVIDIA GeForce RTX 2080 Ti \\
Memory &  32 GB RAM\\
Storage &  1 TB NVMe SSD \\
Operating System & Ubuntu 20.04 LTS \\
\hline
\end{tabular}
\end{table}

\begin{table}[!h]
\caption{Software \& Wireless Configurations}
\label{tab:configurations}
\centering
\begin{tabular}{llp{0pt}ll}
\multicolumn{2}{c}{\textbf{Software Settings}} && \multicolumn{2}{c}{\textbf{Wireless Communication}}\\
\cmidrule{1-2} \cmidrule{4-5}
\textbf{Software} & \textbf{Version} && \textbf{Parameter} & \textbf{Value}\\
\cmidrule{1-2} \cmidrule{4-5}
ROS & Noetic && Carrier frequency & 2.14 GHz\\
Gazebo & 11 && Bandwidth & 20 MHz\\
Sionna RT &  1.2.1 && Subcarriers & 16\\
Mitsuba & 3 && Base station & 3 arrays of \\
Blender & 4.3.2  && & $4\times2$ antennas \\
Python & 3.10 && Input & CSI, PI, \\
PyTorch & 2.4.0 (cu121) && & Spectrogram\\
CUDA & 12.8 && Horizon ($H$) & 5\\
\cmidrule{1-2} \cmidrule{4-5}
\end{tabular}
\end{table}

\begin{table}[H]
\caption{Hyperparameters}
\label{tab:hyper}
\centering
\begin{tabular}{lc c l c}
\midrule
\multicolumn{5}{c}{\textbf{Control JEPA}} \\
\midrule
\textbf{Parameter} & \textbf{Value} && \textbf{Parameter} & \textbf{Value} \\
\midrule
Learning rate & $2\times10^{-4}$ && Discount factor ($\gamma$) & 0.99\\
Batch size & 32 && KL loss scale ($\beta$) & 0.5\\
Gradient clipping & 100 && KL balancing ($\mu$) & 0.8\\
Latent ($z_t$) dim.  & 32  && Replay memory size & $10^{6}$\\
\midrule
\multicolumn{2}{c}{\textbf{Wireless JEPA}} && \multicolumn{2}{c}{\textbf{RL}} \\
\cmidrule{1-2} \cmidrule{4-5}
\textbf{Parameter} & \textbf{Value} && \textbf{Parameter} & \textbf{Value}\\
\cmidrule{1-2} \cmidrule{4-5}
Learning rate & $5\times10^{-3}$ && Actor learning rate & $4\times10^{-5}$\\
Rate decay & 0.97 && Critic learning rate & $10^{-4}$\\
Batch size & 100 && Return weight ($\lambda$) & 0.95\\
EMA decay & 0.99 && Entropy scale ($\eta$) & $10^{-3}$\\
Weight decay & $3\times10^{-3}$ && Slow target update & 1500 \\
\cmidrule{1-2} \cmidrule{4-5}
\end{tabular}
\end{table}

\end{document}

%% file: my_acronyms.tex
\usepackage[acronym]{glossaries}

\newacronym{wm}{WM}{world model}
\newacronym{ros}{ROS}{Robot Operating System}
\newacronym{iiot}{IIOT}{Industrial Internet of Things}
\newacronym{jepa}{JEPA}{Joint Embedding Predictive Architecture}
\newacronym{rf}{RF}{radio frequency}
\newacronym{pid}{PID}{Proportional Integral Derivative}
\newacronym{dqn}{DQN}{Deep Q-Network}
\newacronym{ai}{AI}{Artificial Intelligence}
\newacronym{rl}{RL}{Reinforcement Learning}
\newacronym{csi}{CSI}{Channel State Information}
\newacronym{rssm}{RSSM}{Recurrent State Space Model}
\newacronym{hsv}{HSV}{Hue, Saturation, and Value}
\newacronym{vae}{VAE}{Variational Autoencoder}
\newacronym{rt}{RT}{Ray Tracing}
\newacronym{sdf}{SDF}{Simulation Description Format}
\newacronym{xml}{XML}{Extensible Markup Language}
\newacronym{urdf}{URDF}{Unified Robot Description Format}
\newacronym{rgb}{RGB}{Red, Green, Blue}
\newacronym{vit}{ViT}{Vision Transformer}
\newacronym{cdf}{CDF}{cumulative distribution function}
\newacronym{stft}{STFT}{short-time Fourier transform}
\newacronym{gpu}{GPU}{graphics processing unit}
\newacronym{dt}{DT}{Digital Twin}
\newacronym{itu}{ITU}{International Telecommunication Union}
\newacronym{los}{LoS}{Line-of-Sight}
\newacronym{cjepa}{C-JEPA}{Control Joint Embedding Predictive Architecture}
\newacronym{wjepa}{W-JEPA}{Wireless Joint Embedding Predictive Architecture}
\newacronym{tda}{TDA}{Topological Data Analysis}
\newacronym{pi}{PI}{Persistence Image}
\newacronym{kl}{KL}{Kullback--Leibler}
\newacronym{ap}{AP}{Access Point}

%% file: my_definitions.tex
\newcommand{\sps}[2]{\hl{#1}\textcolor{magenta}{[#2]}}
\newcommand{\tred}[1]{\textcolor{red}{#1}}

\newif\ifshowInstructions

\newcommand{\instructions}[1]{%
	\ifshowInstructions
	\textcolor{blue}{#1}%
	\fi
}

\newcommand{\state}{\vect{x}}
\newcommand{\STATE}{X} 
\newcommand{\stateSpace}{\set{\STATE}}

\newcommand{\set}[1]{\mathcal{#1}}
\newcommand{\vect}[1]{\boldsymbol{#1}}